\documentclass[10pt,twocolumn,letterpaper]{article}

\usepackage{times}
\usepackage{epsfig}
\usepackage{graphicx}
\usepackage{amsmath}
\usepackage{amssymb}
\usepackage{booktabs}

\usepackage{color}
\usepackage{cuted}
\usepackage{capt-of}
\usepackage{tablefootnote}
\usepackage[font=small]{caption} %
\usepackage{multirow} %
\usepackage{footnote}
\usepackage{enumitem}
\usepackage{adjustbox}
\usepackage{subcaption}
\usepackage{caption}
\usepackage{pifont}
\usepackage{algorithm}
\usepackage{algorithmic}
\usepackage{xcolor}
\usepackage{wrapfig}
\usepackage{sidecap}
\usepackage{bbm}
\usepackage[subtle]{savetrees}
\usepackage[subtle]{savetrees}
\usepackage[accsupp]{axessibility}  

\definecolor{ForestGreen}{RGB}{34,139,34}
\definecolor{Cerulean}{RGB}{42,82,190}
\definecolor{CornflowerBlue}{RGB}{100,149,237}
\definecolor{Turquoise}{RGB}{48,213,200}
\definecolor{ProcessBlue}{RGB}{0,136,208}
\usepackage{tabularx}

\definecolor{lightgray}{gray}{0.9}
\definecolor{lightblue}{rgb}{0.93,0.95,1.0}
\definecolor{darkgreen}{rgb}{0.0,0.6,0.0}
\definecolor{darkblue}{rgb}{0.0,0.0,0.5}
\definecolor{pinegreen}{rgb}{0.0, 0.47, 0.44}
\definecolor{deepmagenta}{rgb}{0.8, 0.0, 0.8}
\definecolor{amber}{rgb}{1.0, 0.49, 0.0}

\newcommand{\cmark}{\textcolor{darkgreen}{\ding{51}}}
\newcommand{\xmark}{\textcolor{red}{\ding{55}}}

\newcommand{\ignorebig}[1]{}

\newcommand{\minisection}[1]{\noindent{\textbf{#1}.}}
\newcommand{\tabref}[1]{Table~\ref{#1}}

\newcommand{\figgref}[1]{Figure~\ref{#1}}

\newcommand{\tablestyle}[2]{\setlength{\tabcolsep}{#1}\renewcommand{\arraystretch}{#2}\centering\footnotesize}
\newlength\savewidth

\newcommand{\model}{PromptonomyViT}
\newcommand{\smodel}{PViT}
\newcommand{\gcol}[1]{{\bf \fontsize{6.5}{42}\selectfont \color{citecolor!80}~(#1)}}

\definecolor{citecolor}{RGB}{34,139,34}
\definecolor{lightred}{RGB}{241,140,142}
\definecolor{amber(sae/ece)}{rgb}{1.0, 0.49, 0.0}
\definecolor{battleshipgrey}{rgb}{0.52, 0.52, 0.51}
\definecolor{cadmiumorange}{rgb}{0.93, 0.53, 0.18}

\def\Secref#1{Section~\ref{#1}}

\newcommand\update[1]{\textcolor{red}{[#1]}}

\usepackage{wacv}              

\usepackage{graphicx}
\usepackage{amsmath}
\usepackage{amssymb}
\usepackage{booktabs}

\usepackage[pagebackref,breaklinks,colorlinks]{hyperref}

\usepackage[capitalize]{cleveref}
\crefname{section}{Sec.}{Secs.}
\Crefname{section}{Section}{Sections}
\Crefname{table}{Table}{Tables}
\crefname{table}{Tab.}{Tabs.}

\def\wacvPaperID{1524} 
\def\confName{WACV}
\def\confYear{2024}

\usepackage{amsmath,amsfonts,bm}

\newcommand{\ignore}[1]{}

\def\Secref#1{Section~\ref{#1}}

\def\eqref#1{equation~\ref{#1}}

\def\1{\bm{1}}

\DeclareMathAlphabet{\mathsfit}{\encodingdefault}{\sfdefault}{m}{sl}
\SetMathAlphabet{\mathsfit}{bold}{\encodingdefault}{\sfdefault}{bx}{n}

\begin{document}

\title{PromptonomyViT: Multi-Task Prompt Learning Improves Video Transformers using Synthetic Scene Data}


\author{
    Roei Herzig\thanks{Equal contribution. The order of authors is determined by a coin flip.}~~$^{1,4}$,
    Ofir Abramovich$^*$$^{2}$,
    Elad Ben Avraham$^{1}$,\\
    Assaf Arbelle$^{4}$,
    Leonid Karlinsky$^{5}$,
    Ariel Shamir$^{2}$,
    Trevor Darrell$^{3}$,
    Amir Globerson$^{1}$\\
    \\
    \tt\small
    $^{1}$Tel-Aviv University,
    $^{2}$Reichman University,
    $^{3}$UC Berkeley,
    $^{4}$IBM Research,
    \tt\small
    $^{5}$MIT-IBM Watson AI Lab
}

\maketitle

\begin{strip}
    \centering
    \includegraphics[width=.9\linewidth]{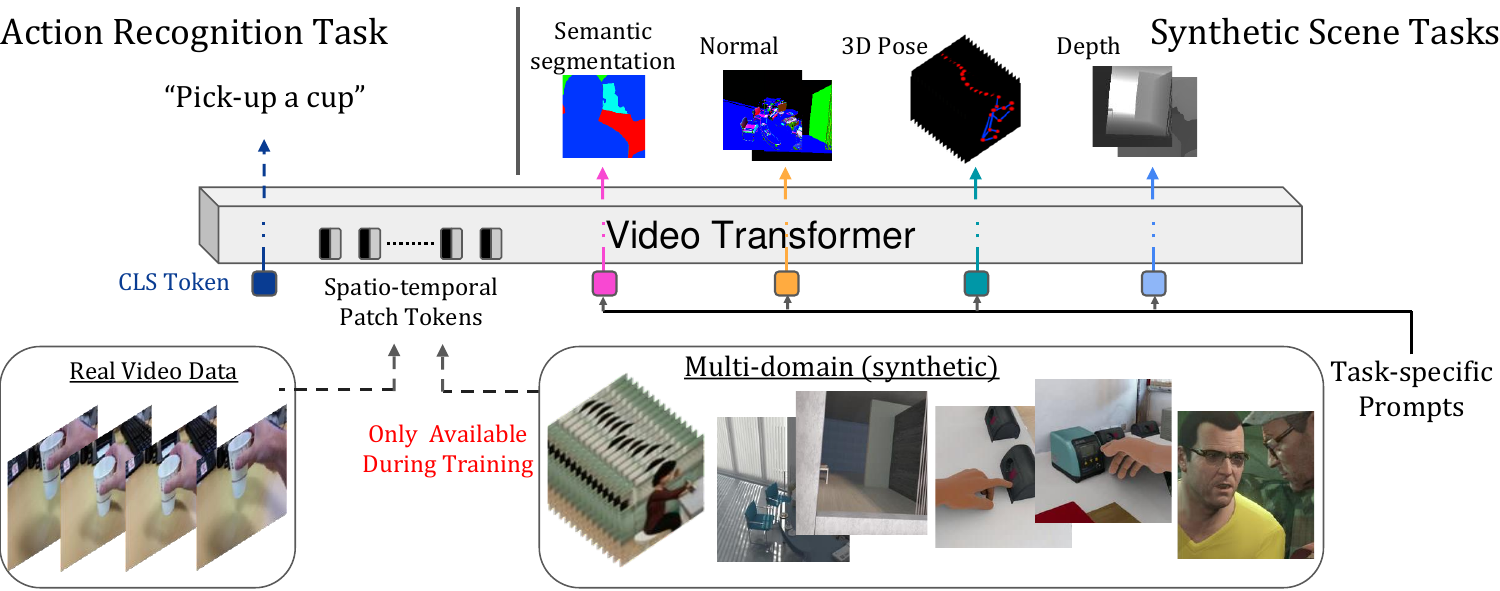}
    \vspace{-0.15cm}
    \captionof{figure}{Our PromptonomyViT ({\smodel}) adds a set of multiple prompts to a video transformer to capture inter-task structure and solve a downstream task. We consider the setting where automatically generated synthetic scene data for scene-level tasks (e.g., depth, semantic segmentation) is used for improving an action recognition model on real data. Our {\smodel} model utilizes a multi-task prompt learning approach for video transformers, where a shared transformer backbone is enhanced with task-specific prompts (colored squares). The task prompts predict the synthetic labels for each task, and a CLS token (blue square) is used to predict the action recognition label. The use of task-specific prompts allows the model to benefit from task-related information.}
\label{fig:teaser}
\end{strip}

\begin{abstract}
    \vspace{-5mm}
    Action recognition models have achieved impressive results by incorporating scene-level annotations, such as objects, their relations, 3D structure, and more. However, obtaining annotations of scene structure for videos requires a significant amount of effort to gather and annotate, making these methods expensive to train. In contrast, synthetic datasets generated by graphics engines provide powerful alternatives for generating scene-level annotations across multiple tasks. In this work, we propose an approach to leverage synthetic scene data for improving video understanding. We present a multi-task prompt learning approach for video transformers, where a shared video transformer backbone is enhanced by a small set of specialized parameters for each task. Specifically, we add a set of ``task prompts'', each corresponding to a different task, and let each prompt predict task-related annotations. This design allows the model to capture information shared among synthetic scene tasks as well as information shared between synthetic scene tasks and a real video downstream task throughout the entire network. We refer to this approach as ``Promptonomy'', since the prompts model task-related structure. We propose the PromptonomyViT model (PViT), a video transformer that incorporates various types of scene-level information from synthetic data using the ``Promptonomy'' approach. PViT shows strong performance improvements on multiple video understanding tasks and datasets. Project page: \url{https://ofir1080.github.io/PromptonomyViT}


\end{abstract}


\section{Introduction}
\label{sec:intro}

Video understanding is a key challenge for machine vision and artificial intelligence. It is intuitively clear that video models should benefit from incorporating spatio-temporal scene-level information including objects, their relations, sizes of instances, 3D structure of a scene, its layout, depth and more. Indeed, several recent studies have explored the use of scene-level information for a variety of video tasks, such as action recognition~\cite{Choutas2018PoTionPM,du2015hierarchical,liu2017enhanced, herzig2022orvit}, action detection~\cite{Kalogeiton2017ActionTD,Zhao2022TubeRTT}, 3D understanding~\cite{Pumarola2021DNeRFNR,Chen20214DContrastCL,Armeni2017Joint2D}, and structured representations for videos~\cite{arnab2021unified,girdhar2019video,Girdhar2017ActionVLADLS,herzig2022orvit,herzig2019stag,ji2019action,Wang_videogcnECCV2018,lvu2021}. However, collecting and annotating real large-scale video datasets~\cite{Ego4D2021,kay2017kinetics} requires an extensive amount of effort and a large budget. This is especially true for complex labels such as 3D structure and segmentation maps.

In the absence of real-world data, synthetic datasets generated by graphics engines~\cite{Savva2019HabitatAP,Gan2021ThreeDWorldAP} provide a powerful alternative for automatically generating scene-level annotations. Graphics engines can be used to generate a large amount of various types of labeled examples of scene-level information. However, learning from synthetic data requires models that can capture those aspects of the synthetic data that are relevant for downstream tasks, and overcome domain gap issues. An additional challenge is how to benefit from multiple types of scene labels (e.g., depth, normal, segmentation maps, 3D joints positions, and more). In this work, we propose a novel approach that can utilize synthetic data of various sources with multiple types of scene annotations to enhance video understanding models. 

Our approach employs Vision Transformers (ViT)~\cite{dosovitskiy2020vit}, which have recently emerged as the leading model for many vision applications~\cite{arnab2021vivit,mvit2021,detr2020}, including for video understanding~\cite{avraham2022svit,herzig2022orvit,li2021improvedmvit,Wu2022MeMViTMM}. Our key insight is that ViT can be naturally extended to multiple synthetic sources through the use of prompt learning. The key idea of prompt learning methods is to augment the transformer input with a set of additional learnable parameters. The notion of prompt learning has been used successfully in NLP~\cite{Lester2021ThePO}, and more recently in machine vision~\cite{Zhou2022LearningTP,Zhou2022ConditionalPL}. Inspired by this, we present a prompt learning approach for video transformers, where a shared backbone is enhanced by a small set of specialized parameters for each task. More specifically, we add a set of ``task prompts'', each dedicated to a unique task. With this design, it is possible to capture information shared among synthetic tasks as well as information shared between synthetic tasks and a real video downstream task, even without applying any domain gap techniques.\footnote{Such techniques may improve performance further, but are orthogonal to our approach.}

The ``task prompts'' construction can be viewed as implementing ``streams of information'', each stream representing a task. This facilitates incorporating information from other tasks into the downstream task, starting from early layers and propagating into the spatio-temporal representations throughout the network. We refer to our prompt-per-task approach as ``Promptonomy'' since the prompts are intended to manage multiple tasks and capture inter-task structure, and name our model {\model} ({\smodel}).\footnote{The name also refers to the classic work on Taskonomy~\cite{Zamir2018TaskonomyDT}, which studied the structure and management of multiple tasks in images.} See Figure~\ref{fig:teaser} for an overview.

Recently, the general idea of prompt tuning has been adapted to vision models by VPT~\cite{Jia2022VisualPT}, suggesting better efficiency of large vision models. Our model differs from recent ``prompt tuning'' approaches in that we refine a full transformer model rather than optimize a limited set of prompt tokens. As a result, information is propagated from the ``task tokens'' to all other tokens, enabling interaction across the entire network between the synthetic tasks and the real video downstream task. Furthermore, our multi-task prompts are supervised by auxiliary tasks, and not the primary action recognition task. 

To summarize, our main contributions are as follows: (i) we propose a new method for exploiting synthetically generated labels for several tasks to improve video understanding models; (ii) we propose the concept of special ``multi-task prompts'' to capture task-related information through task supervision, while also interacting with prompts of other tasks and the downstream video task; (iii) we demonstrate improved performance on five tasks and five datasets on video understanding benchmarks: compositional and few-shot action recognition on SomethingElse, spatio-temporal action detection on AVA, standard action recognition on Something-Something V2, Diving48, and PNR Temporal Localization task on Ego4D, highlighting the effectiveness of the proposed approach.


\section{Related Work}
\label{sec:related}

\minisection{Prompt Tuning} Natural language prompting is a method of reformatting NLP tasks as natural language responses to natural language input. Recently, the concept of prompt tuning for efficient fine-tuning of language models was introduced by~\cite{Lester2021ThePO}. Several recent works~\cite{asai2022attempt, sanh2022multitask,vu-etal-2022-spot}, have explored prompt tuning in the context of multi-task learning in natural language processing. ATTEMPT~\cite{asai2022attempt} suggested a soft prompt tuning approach for parameter efficient multi-task knowledge sharing, UNIFIED PROMPT~\cite{sanh2022multitask} suggested to use multi-task text prompting for zero-shot tasks, and the authors in \cite{vu-etal-2022-spot} suggested the soft prompt tuning method for efficient fine-tuning. Additional recent works~\cite{Jia2022VisualPT,wang2022learning,wang2022dualprompt} suggested exploring the usage of prompt tuning in vision transformers. Specifically, VPT~\cite{Jia2022VisualPT} uses prompt tuning to efficiently fine tune vision transformers, while others~\cite{wang2022learning,wang2022dualprompt} use prompts for continual learning. As opposed to these works, our focus is on the addition of multiple prompts that incorporate various types of scene-level information learned from synthetic data, which will lead to better video understanding. Last, we note that, since our focus is not on efficiency, the entire model is fine-tuned without freezing any parameters.

\minisection{Learning from Synthetic Data} In the field of computer vision, synthetic data has been widely used as an alternative to real-world training data to solve various problems~\cite{Gan2021ThreeDWorldAP,Mikami2021ASL,Prakash2019StructuredDR,Ros2016TheSD,Souza2017ProceduralGO,Savva2019HabitatAP,Varol2021SyntheticHF}. Many works attempted to generate synthetic data that mimics real data for image classification~\cite{Gan2021ThreeDWorldAP,Mikami2021ASL}, semantic segmentation~\cite{Ros2016TheSD,Wang2020DifferentialTF}, action recognition~\cite{Souza2017ProceduralGO,Varol2021SyntheticHF}, object detection~\cite{Peng2015LearningDO,Prakash2019StructuredDR}, representation learning~\cite{Kim2022HowTA, Mishra2022Task2SimTE}, and more~\cite{Savva2019HabitatAP,Xia2018GibsonER}. Instead, our approach focuses on learning multiple tasks simultaneously from several synthetic domains and then transferring knowledge into the real world task by developing a multi-task prompting model and training scheme.

\minisection{Multi-task Learning from Synthetic Data} The multi-task setting refers to the ability to learn multiple tasks simultaneously in which all model parameters are subject to a shared influence~\cite{Caruana1998MultitaskL,Brggemann2020AutomatedSF,Crawshaw2020MultiTaskLW,Kendall2018MultitaskLU,Gao2019NDDRCNNLF,Gao2020MTLNASTN,Sun2020AdaShareLW,Sun2021TaskSN,Liu2021ConflictAverseGD,Liu2021TowardsIM,Yu2020GradientSF,Zhou2020PatternStructureDF}. Many recent works employ multi-task learning in CNNs~\cite{Liu2019EndToEndML,Vandenhende2020MTINetMT} and Transformers~\cite{Xu2022MultiTaskLW,Bhattacharjee2022MuITAE,Hu2021UniTMM} to exploit the potential advantages of fast training, stronger results, and fewer parameters. MTFormer~\cite{Xu2022MultiTaskLW} is a transformer-based architecture, where multiple tasks share the same transformer encoder and decoder but has multiple modules layered on top of that for each task. MuIT~\cite{Bhattacharjee2022MuITAE} is a transformer-based encoder-decoder model with shared attention to learn task inter-dependencies, and UniT~\cite{Hu2021UniTMM} jointly learns multiple tasks across different domains, from object detection to vision-and-language reasoning and natural language understanding. In contrast to these works, our work is a form of prompt-driven auxiliary task learning which uses synthetic scene-level annotations to train video transformers for improving action recognition on real video data. %

\minisection{Scene Understanding Models} Recently, scene understanding models that use scene-level annotations have been successfully applied to a wide range of computer vision applications: panoptic segmentation\cite{Qiao2021ViPDeepLabLV,Cheng2020PanopticDeepLabAS}, video relation understanding~\cite{Liang2019PeekingIT,relnets2017nips,Sun2019VideoVR}, vision and language~\cite{Chen2020UNITERUI,Li2019VisualBERTAS,li2020oscar,Tan2019LXMERTLC}, relational reasoning~\cite{baradel2018object,battaglia2018relational,herzig2018mapping,referential_relationships,Jerbi2020LearningOD,raboh2020dsg,Xu2020reason,zambaldi2018relational}, human-object interactions~\cite{Gao2020DRGDR,Kato2018CompositionalLF,Xu2019LearningTD}, action recognition~\cite{arnab2021unified,girdhar2019video,Girdhar2017ActionVLADLS,herzig2022orvit,herzig2019stag,ji2019action,Nagarajan2020EgoTopoEA,sun2018actor,Wang_videogcnECCV2018,lvu2021,Structured_cvpr19}, and even image \& video generation~\cite{2020ActionGraphs,herzig2019canonical,johnson2018image}. In our work, we demonstrate how video transformers can utilize shared representations from a variety of multiple different synthetic tasks to perform video downstream tasks. 

\minisection{Video Transformers} Vision Transformers~\cite{dosovitskiy2020vit, touvron2021training} recently proposed a new approach to image recognition by discarding the convolutional inductive bias entirely and instead employing self-attention operations. With the advent of ViT, and the fact that attention-based architectures are a natural choice for modeling long-range contextual relationships in video, a number of video transformer models, including TimeSformer~\cite{gberta_2021_ICML}, ViViT~\cite{arnab2021vivit}, Mformer (MF)~\cite{patrick2021keeping}, ORViT~\cite{herzig2022orvit}, MViT~\cite{mvit2021}, MViTv2~\cite{li2021improvedmvit} and Video Swin~\cite{liu2021videoswin}, form the latest era in action recognition. We choose to work with MViTv2, although our method can be used on top of any of these. Our work exploits the seamless ability of the transformer architecture to process multiple domains and to integrate the underlying structure among tasks for several downstream video-related tasks.

\section{The {\smodel} Model}
\label{sec:model}

Our {\smodel} approach utilizes synthetic data of various domains with multiple types of scene annotations to enhance video understanding models. We consider the setting in which the main goal is to learn downstream video-understanding tasks, such as action recognition or action detection, while leveraging multiple synthetic scene-annotated datasets. The key idea of our work is that multi-task prompt learning can be used to incorporate synthetic scene tasks into the video model. This is achieved by adding a set of \emph{task prompts}, each corresponding to a different task, and letting each prompt predict task-related annotations. Importantly, all prompts are part of the computation for any video, regardless of the underlying task, and thus enables sharing information among auxiliary tasks.

\begin{figure*}[t!]
    \centering
    \includegraphics[width=.9\linewidth]{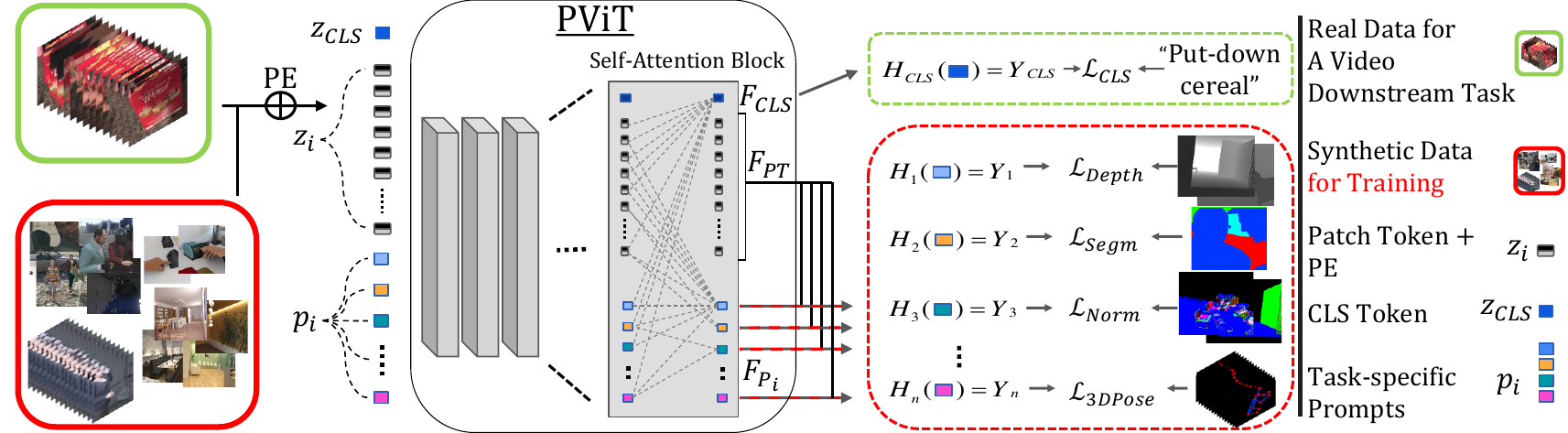}
    \caption{\textbf{{\smodel} architecture}. We extend a transformer with a set of ``task prompts'', $p_i$, that are designed to capture information regarding each task, as well as capture the inter-task structure. The prompts are supervised by synthetic scene auxiliary tasks (depth, segmentation, normal, and 3D pose) available only during training, in order to enhance performance on a video task (predicting ``put-down cereal''). Each task prompt in the attention block interacts with the patch tokens and CLS token, as well as other task prompts within the block.}
    \label{fig:arch}
    \vspace{-1.1em}
\end{figure*}

We begin by describing the video transformer architecture and the training setup (\Secref{sec:model:preliminaries}). We then introduce our Multi-task Prompts (\Secref{sec:model:task_tokens}) and the Training losses (\Secref{sec:model:losses}). Our method is illustrated in~\figgref{fig:arch}.

\vspace{-0.1em}
\subsection{Preliminaries}
\label{sec:model:preliminaries}
\vspace{-0.1em}

\minisection{Video Transformer Architecture}
A typical Video Transformer model takes as input a video $X\in\mathbb{R}^{T \times 3 \times H \times W}$, extracts $N$ non-overlapping per-frame patches $x_{i}\in\mathbb{R}^{3\times h\times w}$ and projects them into a lower-dimension $d$ (e.g., see \cite{dosovitskiy2020vit}). Denote the transformer patches by $Ex_i$, which we refer to as ``patch tokens''. Then, spatio-temporal position embeddings $\text{PE} \in \mathbb{R}^{N \times d}$ are added for providing location and time location information, resulting in a new embedding: $z_{i} = Ex_{i} + \text{PE}_{i}$. This forms the sequence of input tokens to the video transformer:
\begin{equation}
    z = [z_{CLS},z_1, z_2,\cdots,z_{N}]
\end{equation}
where $z_{CLS}$ is a CLS token used for the downstream task. Next, a transformer is comprised of a stack the Multi-headed Attention (MHSA) blocks, which apply the self-attention operation over all patch tokens $z$ (including the CLS token $z_{CLS}$) followed by a Feed-Forward Network (FFN), a layer normalization (LayerNorm~\cite{Ba2016LayerN}) step and a non-linear operation with residual connections~\cite{resnet1}.

\minisection{Training Setup for Various Domains}
In our approach, we aim to process batches of videos from various domains for $n$ different tasks. A key desideratum in this context is to be able to input both videos of synthetic scene data across various domains for multiple tasks, as well as videos from the real domain into the same model. In contrast to standard training, where each sample contains a full set of annotations (e.g., depth, normal, etc.), in our case, only partial annotations are included. This is explained in greater detail in~\Secref{sec:model:training}. 

\subsection{Multi-task Prompts}
\label{sec:model:task_tokens}
As mentioned earlier, our key observation is that multi-task prompt learning can be used to incorporate synthetic scene tasks into the video model. Towards this end, we add a set of ``task prompts'' designed to capture information regarding each task, as well as capture the inter-task structure. Specifically, we define a fixed number of $n$ learned vectors $p_1,p_2,\cdots,p_{n}\in \mathbb{R}^{1\times d}$ for tasks $T_1,\cdots,T_{n}$. We refer to these vectors as the learned task prompts.

Let $P = \{p_1,p_2,\cdots,p_{n}\}$ be the set of task prompts. These prompts are concatenated to the patch tokens to obtain the following set of inputs to the transformer:
\begin{equation}
    z = \left[z_{CLS},z_1,z_2,...,z_N,p_1,p_2,...,p_n\right] %
\end{equation}

The transformer processes the input $z$, resulting in a new representation for each token $z$ (i.e., the CLS token, the patch tokens, and the task-prompts). We denote $F_{CLS}(z)$ as the representation of the CLS token, and let $F_{P_i}(z)$ denote the representation of the $i^{th}$ task-prompt. We also use $F_{PT}(z)$ as the final representation of all the patch tokens.

\begin{figure}[t!]
    \centering
    \includegraphics[width=1\linewidth]{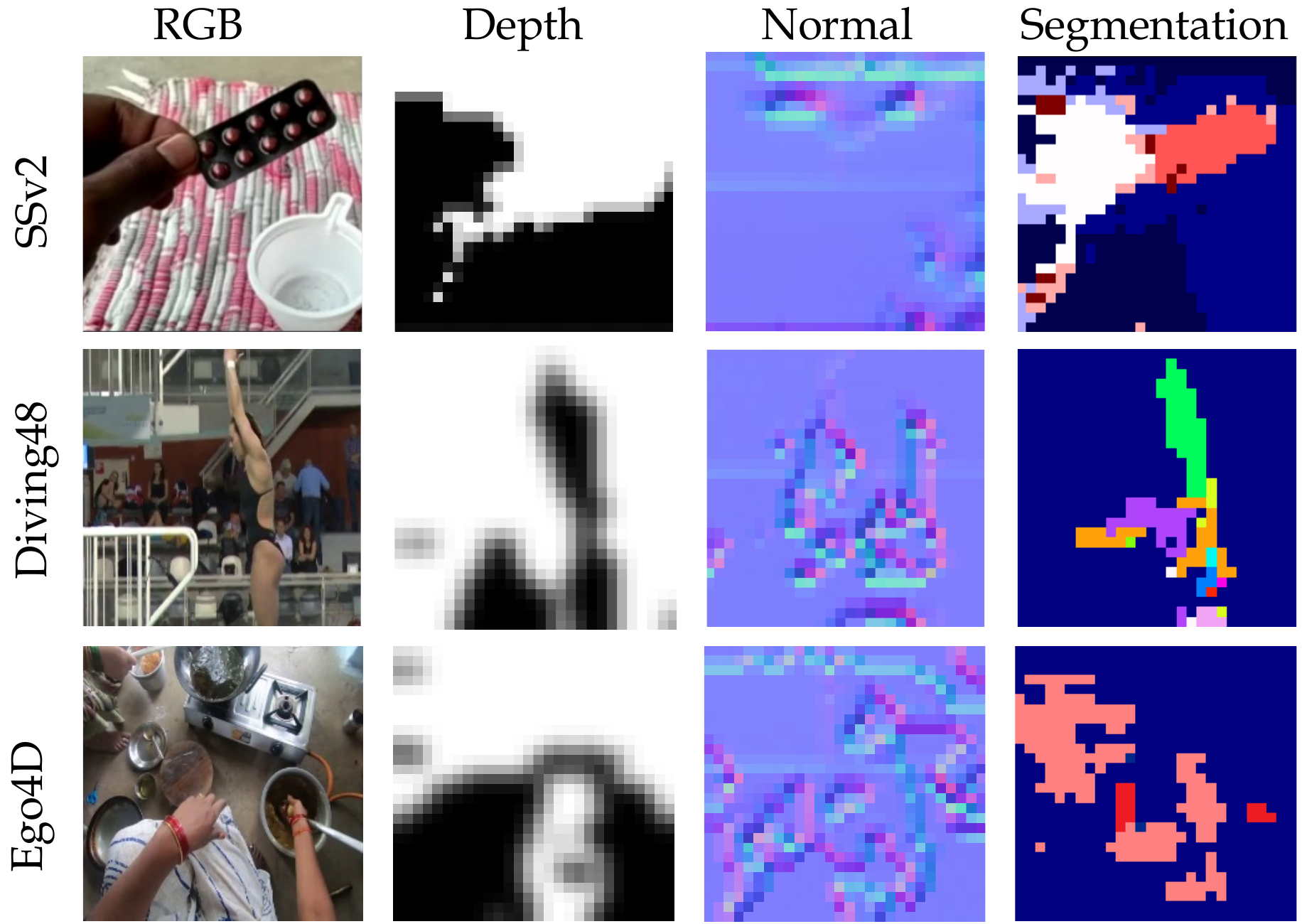}
    \captionof{figure}{\textbf{``Task Prompts'' Visualization}. Visualization of the output of the ``task prompts'' prediction heads on frames from the SSv2, Diving48, and Ego4D datasets. The model was trained with Something-Else as the action recognition dataset. Shown are prediction head outputs (i.e., $H_i$) for depth, normal, and semantic segmentation. It can be seen that the task prompts produce meaningful maps, despite not receiving such labels for real videos.
    }
    
    \label{fig:vis_objs}
    \vspace{-1.em}
\end{figure}

Next, these final output tokens are used for predicting labels. For the action recognition task, we simply predict using  $F_{CLS}(z)$ and a prediction head $\hat{Y}_{CLS} = H_{CLS}(F_{CLS}(z))$. For the synthetic tasks, the task $i$ has a prediction head $\hat{Y_{i}} = H_{i}(F_{P_i}(z), F_{PT}(z))$ that is used for predicting labels corresponding to this task. It uses the patch tokens only for cases where a dense prediction is required (e.g., segmentation maps, normal and depth estimation). The task heads $H_i$ for localization tasks (e.g., boxes and 3D poses), are a simple FC layer, while for dense prediction tasks, we upsample patch token outputs from several layers and concatenate them with the corresponding task token to predict the task output map. \figgref{fig:vis_objs} also visualizes the ``task prompts'' learned by our model. For more info about the prediction heads see~\Secref{supp:impl} in Supplementary.

\subsection{Training and Inference}
\label{sec:model:training}

Our training data consists of labeled examples from $n$ synthetic tasks, as well as the downstream task of action recognition. As mentioned above, we have $n+1$ predictions heads corresponding to those. During training, for each training video we add a loss corresponding to the labels provided for that video. For example, if the synthetic video $X$ contains labels for task $2$ (e.g., depth) and task $5$ (e.g., normal), we take the output of prediction heads $F_2$ and $F_5$ and compare them to the ground-truth labels for these two tasks.  We formally describe the task-specific losses below. We  use $\hat{Y}$ to refer to predicted labels, and $Y$ for ground-truth labels.

\minisection{Losses}
\label{sec:model:losses}
For Depth Estimation, we first downsample the ground-truth depth map $Y_{Depth}$ to a fixed scale of $\tilde{h} \times \tilde{w}$ map. Next, we predict a fixed scale map $\hat{Y}_{depth}$, and clip large values to focus on relatively closer objects. Finally, we use the MSE loss for computing the per-pixel depth error:
\begin{equation}
    \label{loss:depth}
    \mathcal{L}_{Depth} = \frac{1}{\tilde{h} \times \tilde{w}} \cdot \text{MSE}\left({\hat{Y}_{Depth}},Y_{Depth}\right)
\end{equation}

For Normal Estimation, we predict the normal map $\hat{Y}_{Normal} \in \mathbb{R}^{h \times w \times 3}$ for every axis in world coordinates. We again down-sample the ground truth map $Y_{Normal}$ and compute the MSE loss:
\begin{equation}
    \label{loss:normal}
    \mathcal{L}_{Normal} = \frac{1}{\tilde{h} \times \tilde{w}} \cdot \text{MSE}\left(\hat{Y}_{Normal},Y_{Normal}\right)
\end{equation}

For Semantic Segmentation, we use per-pixel multi-label classification to compute a map for different semantic instances in the scene. We downsample the ground-truth map ${Y}_{Segm}$ and compute pixel-level cross-entropy loss followed by a Softmax function:
\begin{equation}
    \mathcal{L}_{Segm} = \frac{1}{\tilde{h} \times \tilde{w}} \cdot \text{CE}\left(\hat{Y}_{Segm}, Y_{Segm}\right)
\end{equation}

For 3D Pose estimation, we predict a tensor $\hat{Y}_{Pose3d} \in \mathbb{R}^{1 \times 75}$ corresponding to a $25\times3$ of 3D joins in KinectV2 format~\cite{CARUSO2017174}. Each training sample consists of a single individual. We define the loss for 3D Pose Estimation to be: 
\begin{equation}
    \mathcal{L}_{3DPose} = \frac{1}{75} \cdot \text{MSE}\left(\hat{Y}_{3DPose},Y_{3DPose}\right)
\end{equation}

For Bounding Box Prediction, we set a fixed number of $O$ objects per training sample and use the $L1$ loss function to compute boxes predictions $\hat{Y}_{Boxes} \in \mathbb{R}^{O \times 4}$ and the corresponding ground-truth coordinates $Y_{Boxes}$:
\begin{equation}
    \mathcal{L}_{Boxes} = \text{L}_{1}\left(\hat{Y}_{Boxes},Y_{Boxes}\right) + \text{GIoU}\left(\hat{Y}_{Boxes},Y_{Boxes}\right)
\end{equation}
where the GIoU is used as in~\cite{Rezatofighi_2018_CVPR}.

Last, for the video downstream task (denoted as $DT$), on which we evaluate our model, we consider the standard cross-entropy loss between the predicted logits $\hat{Y}_{CLS}$ and the true video labels $Y_{CLS}$ as follows:
\begin{equation}
    \mathcal{L}_{DT} = \text{CE}\left(\hat{Y}_{CLS},Y_{CLS}\right)
\end{equation}

The total loss is the sum of all of the losses described above. We note that only losses for which the samples have ground truth are added since the ground truth changes across instances, as our training does not use explicit correspondences between different input modalities. Each of the task terms in the loss is multiplied by a hyper-parameter ($\lambda$), and these were chosen such that all loss components have the same scale (see Supplementary).
The total loss is the weighted combination of all terms:
\begin{equation}
\label{sec:eq:all_losses}
\begin{matrix}
    \mathcal{L}_{Total} = \lambda_{DT}\mathcal{L}_{DT} + \lambda_{Depth}\mathcal{L}_{Depth} + \lambda_{Normal}\mathcal{L}_{Normal} \\ 
    ~~~~~~+\lambda_{Segm}\mathcal{L}_{Segm} +  \lambda_{3DPose}\mathcal{L}_{Pose3d} + \lambda_{Boxes}\mathcal{L}_{Boxes}
\end{matrix}
\end{equation}

For simplicity, we omit the temporal dimension when predicting the losses above per frame.

\minisection{Inference} For inference, {\smodel} receives input from the real videos without requiring any additional synthetic data. 

Finally, our method can be applied on top of a variety of video transformers (MViT~\cite{mvit2021}, TimeSformer~\cite{gberta_2021_ICML}, Mformer~\cite{patrick2021keeping}). For our experiments, we use the MViTv2~\cite{li2021improvedmvit} model because it performs well empirically.

\section{Experiments and Results}
\label{sec:expr}

We begin by describing the datasets (\Secref{sec:expr:datasets}), implementation details (\Secref{sec:expr:impl}), and baselines (\Secref{sec:expr:baselines}). Next, we evaluate our approach on several benchmarks and tasks. Specifically, we consider the following tasks: Compositional Action Recognition (\Secref{sec:expr:comp_fewshot}), Object State Change Classification \& Localization (\Secref{sec:expr:ego4d}), Action Recognition (\Secref{sec:expr:ar}), and Spatio-Temporal Action Detection (\Secref{sec:expr:action_detection}).

\subsection{Datasets}
\label{sec:expr:datasets}

We first describe the datasets used for the downstream video tasks, followed by the datasets used as auxiliary synthetic datasets including their annotations. We use the following video datasets: \textbf{(1) Something-Something v2 (SSv2)}~\cite{goyal2017something} is a dataset containing 174 action categories of common human-object interactions. \textbf{(2) SomethingElse~\cite{materzynska2019something}} which exploits the compositional structure of SSv2, where a combination of a verb and a noun defines an action. We follow the official compositional split from~\cite{materzynska2019something}, which assumes the set of noun-verb pairs available for training is disjoint from the set given at test time. \textbf{(3) Ego4D}~\cite{Ego4D2021} is a new large-scale dataset of more than 3,670 hours of video data, capturing the daily-life scenarios of more than 900 unique individuals from nine different countries around the world. \textbf{(4) Diving48}~\cite{Li_2018_Diving48} contains 48 fine-grained categories of diving activities. \textbf{(5) Atomic Visual Actions (AVA)}~\cite{AVA2018} is a benchmark for human action detection, we report Mean Average Precision (mAP) on AVA-V2.2. For ``auxiliary'' synthetic datasets, we use \textbf{(1) SURREACT}~\cite{varol21_surreact}, a novel synthetic data generation method based on real human motion from real datasets. The method renders 3D SMPL~\cite{Loper2015SMPLAS} sequences with randomized cloth textures, lighting, and body shapes from 3D skeleton joints extracted by Kinect V2~\cite{kay2017kinetics} from the two following datasets: \textbf{(i) NTU RGB+D}~\cite{shahroudy2016ntu} is a large-scale multi-view video dataset of RGB-D human actions with 56,880 samples collected from 40 subjects, including depth maps and 3D skeleton joints. \textbf{(ii) UESTC RGB-D}~\cite{ji2019largescale} is also a multi-view action dataset that with 40 categories of aerobic exercise along with depth maps and 3D skeleton joints. \textbf{(2) HyperSim}~\cite{roberts:2021} is a photorealistic synthetic dataset for holistic indoor scene understanding. This dataset contains 77,400 HD images of 461 indoor scenes as well as ground truth depth and normal values for each pixel. \textbf{(3) Procedural Human Action Videos (PHAV)}~\cite{DeSouza:Procedural:CVPR2017} is a human action video dataset which relies on procedural generation and other computer graphics techniques of modern game engines. There are 39,982 actions in 35 categories, annotated with optical flow, segmentation, and depth maps. \textbf{(4) KIST SynADL}~\cite{hwang2020eldersim} generated by the ElderSim engine, is a large-scale synthetic dataset of elders’ activities. There are 462K RGB videos representing 55 action classes, along with 2D, 3D skeleton joints positions used as ground truth. \textbf{(5) EHOI}~\cite{leonardi2022egocentric} consists of 20K synthetic image dataset of first-person view, annotated with segmentation masks, and hand-object interaction boxes of 19 categories.

\begin{table}[t!]
  \vspace{-0.5em}
  \scriptsize
  \tablestyle{2.0pt}{1.00}
   \centering
    \begin{tabular}{l cc cc cc cc}
            \toprule
            
            \multirow{2}{*}{{Model}} & \multicolumn{2}{c}{Compositional} & \multicolumn{2}{c}{Base} & \multicolumn{2}{c}{Few-Shot} \\ 
            
             & {Top-1} & {Top-5} & {Top-1} & {Top-5} & {5-Shot} & {10-Shot}  \\
            
            \midrule
            {I3D~\cite{carreira2017quo}} & 42.8 & 71.3 & 73.6 & 92.2 & 21.8 & 26.7 \\
            {SlowFast~\cite{slowfast2019}} & 45.2 & 73.4 & 76.1 & 93.4 & 22.4 & 29.2 \\
            {TimeSformer~\cite{gberta_2021_ICML}} & 44.2 & 76.8 & 79.5 & 95.6 & 24.6 & 33.8 \\
            {Mformer~\cite{patrick2021keeping}} &  60.2 & 85.8 & 82.8 & 96.2 & 28.9 & 33.8 \\
            {MViTv2~\cite{li2021improvedmvit}} &  63.3 & 87.5 & 83.7 & 96.8 & 32.7 & 40.2 \\
            
            \midrule
            {MViTv2 MT} &  63.0 & 87.6 & 79.8 & 95.8 & 32.7 & 40.6 \\
            {MViTv2 VPT} & 53.0 & 81.8 & 76.8 & 94.8 & 31.8 & 39.0 \\
            \midrule
            \multirow{2}{*}{\textbf{PViT (Ours)}} & \textbf{65.5} & \textbf{89.0} & \textbf{85.0} & \textbf{97.4} & \textbf{34.3} & \textbf{41.3} \\
            & \gcol{$+$2.2} & \gcol{$+$2.5} & \gcol{$+$1.3} & \gcol{$+$0.6} & \gcol{$+$1.6} & \gcol{$+$1.1} \\
            \bottomrule
   \end{tabular}%
    \vspace{-1.0em}
    \caption{\textbf{Compositional and Few-Shot Action Recognition} on the SomethingElse dataset.}
    \label{tab:comp_fewshot}
    \vspace{-1.5em}
\end{table}

\vspace{-0.1cm}
\subsection{Implementation Details}
\label{sec:expr:impl}
\vspace{-0.1cm}

{\smodel} is implemented in PyTorch, and the code will be released upon acceptance and is included in the supplementary. Our training recipes and code are based on the MViTv2-S, $16\times4$ model, and were taken from \url{https://github.com/facebookresearch/mvit}. We pretrain the {\model} model on the K400~\cite{kay2017kinetics} video dataset. Then, we finetune on the downstream video task (detailed in \Secref{sec:expr:datasets}) with the synthetic datasets and the {\model} loss. In the training batch, there are 64 videos with the number of synthetic videos being at most $\times3$ the number of real videos. For more implementation details, see~\Secref{supp:impl} in Supplementary.

\vspace{-.3em}
\subsection{Baselines}
\label{sec:expr:baselines} 
\vspace{-0.3em}

In our experiments, we compare {\smodel} to several models reported in previous work for the corresponding datasets. These include the following methods: BMN~\cite{Lin2019BMNBN}, \textit{I3D}~\cite{carreira2017quo}, \textit{SlowFast}~\cite{slowfast2019}, as well as the state-of-the-art transformers -- \emph{SViT}~\cite{avraham2022svit}, \emph{TimeSformer}~\cite{gberta_2021_ICML}, \emph{ViViT}~\cite{arnab2021vivit}, and \emph{MViTv2}~\cite{li2021improvedmvit}. 

Additionally, we explore two alternative ViT-based baselines. First, we consider a model we call \emph{MViTv2 multi-task (MViTv2 MT)}, and is perhaps the simplest application of ViT to our task. It augments the MViTv2 model with multiple prediction heads (one per synthetic task) operating on the CLS token,  but \textit{does not use additional task prompts}. The prediction heads have the same architecture as $H_i$ used in {\smodel}. We also consider a model we refer to as \emph{MViTv2 VPT}, which is an implementation of the VPT~\cite{Jia2022VisualPT} approach for action recognition. This is a simple prompt-based approach  utilizes the additional task prompts included in PViT but does not use additional synthetic data and \emph{keeps the backbone frozen}. The advantage of MViTv2 VPT is training efficiency, as fewer parameters are used in training. Considering VPT trains only a few parameters, we assume that the parameters are insufficient to account for differences between pretraining (K400) and target datasets (AVA, SSv2). This is in marked difference to the case of images where VPT worked (their pretraining and target benchmarks are similar in distribution and tasks). Nevertheless, we still find it important to evaluate their method in our setting.


\begin{table}[t!]

    \vspace{-0.5em}
	\centering
	\tablestyle{2.0pt}{1.0}
	\scriptsize
	\begin{tabular}{l c c c c}
	    \toprule
		\multirow{2}*{Model} & {Temporal} & {PNR}
		\\ & {Localization Error} & {Classification Top-1}
		\\
		\midrule
		
        {Bi-LSTM}           & 0.790 & 65.3 \\
        {BMN~\cite{Lin2019BMNBN}}               & 0.780 & - \\
        {I3D ResNet-50~\cite{carreira2017quo}}     & 0.739 & 68.7 \\
        {MViTv2~\cite{li2021improvedmvit}}           & 0.702 & 71.6 \\

		\midrule
        {MViTv2 MT}           & 0.640 & 73.6 \\
        {MViTv2 VPT}           & 0.791 & 64.2 \\
        \midrule
		{\textbf{PViT (Ours)}} & \textbf{0.637 \gcol{-0.065}} & \textbf{74.8 \gcol{+3.2}} \\
            \bottomrule
	\end{tabular}
	\vspace{-1.0em}
 	\caption{{\textbf{PNR Temporal Localization} results on Ego4D.}}
	\label{tab:sota_ego}
	\vspace{-1.7em}
\end{table}

\renewcommand{\thefootnote}{\fnsymbol{footnote}}
\begin{table*}[tb!]
    \centering
    \begin{subtable}[t]{.34\linewidth}
	\tablestyle{3pt}{1.25}
	\scriptsize
	    \centering
        \caption{\bf{Something--Something V2}}
	    \label{tab:sota_ssv2}
		\begin{tabularx}
{\linewidth}{@{}l c c c}
    \toprule
    {Model} & Pretrain & {Top-1} & {Top-5}\\ 
    \midrule
    SlowFast~\cite{slowfast2019}, R101 & K400 & 63.1 & 87.6 \\
    ViViT-L~\cite{arnab2021vivit} & {\scriptsize IN+K400}  & 65.4 & 89.8 \\
    {MViTv1~\cite{mvit2021}} & {\scriptsize K400}  &  64.7 &  89.2 \\
    {MViTv2~\cite{li2021improvedmvit}} & {\scriptsize K400}  &  68.2 &  91.4 \\
    \midrule
    {MViTv2 MT} & {\scriptsize K400}  &  68.4 &  91.4 \\
    {MViTv2 VPT} & {\scriptsize K400}  &  61.5 &  87.5 \\
    \midrule

    
    
    

    \bf PViT (Ours) & {\scriptsize K400}  & \textbf{69.4}\gcol{+1.2}  & \textbf{91.6}\gcol{+0.2} \\

    \bottomrule
\end{tabularx}

	\end{subtable}
    \begin{subtable}[t]{.30\linewidth}
    \tablestyle{3pt}{1.25}
    \scriptsize
        \centering
		\caption{\bf{Diving48}}
        \label{tab:sota_diving48}
		\renewcommand*{\thefootnote}{\fnsymbol{footnote}}
\setlength{\tabcolsep}{3pt}

\begin{tabularx}
{\linewidth}{@{}l c c c}
    \toprule
    {Model} & {Pretrain~} & {Frames} & {Top-1} \\ 
    \midrule

    SlowFast~\cite{slowfast2019}, R101 & K400 & 16 & 77.6 \\
    TimeSformer~\cite{gberta_2021_ICML} & IN & 16 & 74.9 \\
    MViTv2~\cite{li2021improvedmvit} & K400 & 16 & 73.1\\
    SViT~\cite{avraham2022svit} & K400 & 16 & 79.8\\
    
    
    \midrule
    {MViTv2 MT} & K400 & 16 & 82.2 \\
    {MViTv2 VPT} & K400 & 16 & 69.8 \\
    \midrule
    \textbf{PViT (Ours)} & K400 & 16 & \textbf{85.8}\gcol{+6.0} \\

    \bottomrule
\end{tabularx}

	\end{subtable}
	\begin{subtable}[t]{.30\linewidth}
	\tablestyle{6pt}{1.25}
	\scriptsize
	    \centering
	    \caption{\bf{AVA-V2.2}}
	    \label{tab:sota_ava}
	\begin{tabular}{l c c c}
	    \toprule
		{Model} & {Pretrain} & {mAP} \\ 
		\midrule
		{SlowFast~\cite{slowfast2019}}, R50 & {K400} & 22.7 \\ 
		{SlowFast~\cite{slowfast2019}}, R101 & {K400} & 23.8 \\ 
		{MViTv1~\cite{mvit2021}}  & {K400} & 25.5 \\ 
		{MViTv2~\cite{li2021improvedmvit}}  & {K400} & 26.8 \\
		\midrule
        {MViTv2 MT}  & {K400} & 26.3 \\
        {MViTv2 VPT}  & {K400} & 19.0 \\
		\midrule
		\textbf{PViT (Ours)} & {K400} & \textbf{28.4}\gcol{+1.6} \\
		\bottomrule
	\end{tabular}

	\end{subtable}
    \caption{\textbf{Results on SSv2, Diving48, and AVA-V2.2 datasets.} We report (a) Top-1 and top-5 accuracy on SSv2. (b) Top-1 on Diving48. (c) mAP metric on AVA. IN refers to ImageNet-21K. For additional comparisons, see~\Secref{supp:expr:baselines} in supplementary.} 
    \vspace{-1.5em}
    \label{table:ar}
\end{table*}
\renewcommand{\thefootnote}{\arabic{footnote}}

\subsection{Compositional \& Few-Shot Action Recognition}
\label{sec:expr:comp_fewshot}

In several video datasets, an action is defined as the combination of a verb and a noun. Hence, one of the challenges is to identify combinations of words that were not seen during training. This ``compositional'' setting was explored in the ``SomethingElse'' dataset~\cite{materzynska2019something}, where verb-noun combinations in the test data do not occur in the training data. We also evaluate the few-shot compositional action recognition task in~\cite{materzynska2019something} (See \Secref{supp:impl:smthelse} in supplementary).

\tabref{tab:comp_fewshot} reports the results for these two tasks. {\smodel} outperforms MViTv2 baseline for both the \textit{Compositional} and \textit{Few-shot} tasks by 2.2\% for the compositional task, and by 1.6\%, 1.1\% for 5 and 10-shot tasks. Furthermore, PViT outperforms MViTv2 MT, suggesting that the design of our task prompts approach is beneficial for learning from synthetic data. It can also be seen that MViT VPT performance is adversely affected, as suggested above, resulting in 53\%.



\subsection{Object State Change Tasks}
\label{sec:expr:ego4d}

Human activity relies heavily on hands and objects. Two tasks studying hand-object interaction have recently been introduced to the Ego4D~\cite{Ego4D2021} dataset. The first is temporal localization, which involves finding key frames that indicate a change in object state within a video clip. The second is the classification of object state changes, which indicates whether an object state has changed or not.

\tabref{tab:sota_ego} reports results on the above two tasks in Ego4D. We observe that {\smodel} performs better than MViTv2 by 3.2\%/-0.065 on the classification/localization tasks. As in~\Secref{sec:expr:comp_fewshot}, it can be seen that {\smodel} consistently outperforms MViTv2 MT and MViTv2 VPT baselines. Overall, these results indicate that {\smodel} successfully leverages scene data, even for another downstream video task. 

\subsection{Action Recognition}
\label{sec:expr:ar}

Tables \ref{tab:sota_ssv2} and \ref{tab:sota_diving48} report results for the standard action recognition task on the SSv2 and Diving48 datasets. It can be seen that in Diving48, our method improves over the MViTv2 baseline by 6.0\%, outperforming the other methods.
We hypothesize that this relatively high gain is due to (i) the large availability of synthetic pose annotations (which is likely to help in human actions in the Diving dataset; See \figgref{fig:dataset-task}). (ii) Since Diving is a small dataset, the introduction of additional synthetic supervision results in a larger effect. Finally, {\smodel} achieves a 1.2\%, improvement in SSv2, indicating that {\smodel} can improve on large datasets (180K videos). Last, {\smodel} consistently outperforms MViTv2 MT and MViTv2 VPT baselines, as above. 

\subsection{Spatio-temporal Action Detection}
\label{sec:expr:action_detection}

Gu et al.~\cite{AVA2018} describes the action detection task on AVA as a two-stage prediction procedure. As a first step, boxes are detected using an off-the-shelf person detector, followed by a prediction of the action of each detected box. For fair comparisons, the person boxes are kept identical across approaches, and the final result is measured by the Mean Average Precision (MAP) metric.

\tabref{tab:sota_ava} reports results for spatio-temporal action detection on the AVA dataset. We observe that {\smodel} improves the MViTv2 baseline by 1.6\%, thereby demonstrating the ability to leverage ``task prompts'' to detect and localize human actions. In addition, {\smodel} consistently outperforms MViTv2 MT and MViTv2 VPT baselines, as above.

\ignore{
\subsection{Alternative  ViT Approaches}
\label{sec:expr:vit_approaches}
\update{
The results in Tables 1-3 indicate that MViTv2 MT underperforms {\smodel} with gaps between 1.0\% and 3.6\%, supporting the use of task prompts. Last, we note that MViT VPT performance is adversely affected in all tables 1-3 since the number of training parameters is only 0.3\% of the MViTv2 baseline (See~\tabref{tab:abl_pvit}). Based on these results, it can be concluded that the VPT alternative may not be the optimal approach for improving video transformers.}
}

\subsection{Ablations}
\label{sec:ablations}

We perform a comprehensive ablation study on the ``SomethingElse''~\cite{materzynska2019something} dataset to measure the contribution of the different {\smodel} components (See \tabref{tab:ablations}). For more ablations, see~\Secref{supp:expr} in supplementary.

\setlength{\tabcolsep}{2pt}

\begin{table*}[tb!]
    \begin{tabular}{ c c }
        \begin{subtable}[t]{.51\linewidth}
            \tablestyle{4.2pt}{1.0}
            \scriptsize
            \centering
            \caption{\bf{The Role of Prompts and Tuning}}
            \label{tab:abl_pvit}
%

\begin{tabularx}
{\linewidth}{@{}l c c c c c c}
    \toprule
    ~~\multirow{2}{*}{Model} & ~~\multirow{2}{*}{Top-1} & ~~\multirow{2}{*}{Top-5} & {Synthetic} & {Train/Test} & {FLOPS} & {Runtime}\\ 
    ~~{} & {} & {} & {Data} & {Params ($\times10^6$)} & {($\times10^6$)} & {(ms)}\\ 
    \midrule
    ~~{MViTv2~\cite{li2021improvedmvit}}      & 63.3 & 87.5 & \xmark & 38.2/38.2 & 70.6
 & 132.2\\
    ~~MViTv2 MT   & 62.7 & 87.6 & \cmark & 45.0/38.2 & 89.3 & 131.4 \\
    ~~MViTv2 OP   & 63.5 & 88.0 & \cmark & 45.0/38.2 & 89.5 & 137.7 \\
    ~~MViTv2 NP   & 63.4 & 87.8 & \xmark & 38.2/38.2 & 79.9 & 154.5\\ 
    \midrule
    ~~MViTv2 VPT   & 53.0 & 81.8 & \xmark & 0.13/38.2 & 82.3 & 154.5 \\
    ~~PViT VPT   & 53.9 & 82.4 & \cmark & 7.2/38.2 & 93.9 & 143.7 \\
    \midrule
    ~~PViT   & \textbf{65.5} & \textbf{89.0} & \cmark & 45.0/38.2 & 93.9 & 142.8 \\

    \bottomrule
\end{tabularx}

        \end{subtable}
        
    &
        \begin{subtable}[t]{.5\linewidth}
            \caption{\bf{Effect of Synthetic Data Size}}
            \centering
            \scriptsize
            
            \includegraphics[width=1.\linewidth, height=.25\linewidth]{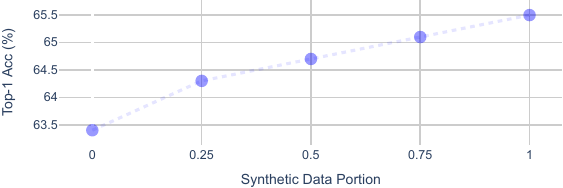}
            \label{fig:ratios}
        \end{subtable}
    \\

        \begin{subtable}[t]{.5\linewidth}
            \tablestyle{4.5pt}{1.0}
            \scriptsize
            \centering
            \caption{\bf{Auxiliary Tasks Contribution}}
            \label{tab:abl_datatasks}

\setlength{\tabcolsep}{3pt}

\begin{tabularx}
{\linewidth}{@{}l c c c c c c c c r@{} r}
    \toprule
    ~~{Datasets} & {Depth} & {Segm.} & {Normal} & {3D Poses} & {2D Boxes} & {Top-1} & {Top-5} \\
    \midrule
    ~~-      & \xmark & \xmark & \xmark & \xmark & \xmark & 63.3 & 87.5 \\
    \midrule
    ~~PHAV+HS+SURR  & \cmark & \xmark & \xmark & \xmark & \xmark & 64.8 & 88.7 \\
    ~~SUR+EHOI  & \xmark & \cmark & \xmark & \xmark & \xmark & 65.0 & 88.7 \\
    ~~HS  & \xmark & \xmark & \cmark & \xmark & \xmark & 63.9 & 88.2 \\
    ~~SUR+ES  & \xmark & \xmark & \xmark & \cmark & \xmark & 64.1 & 88.4 \\
    ~~EHOI      & \xmark & \xmark & \xmark & \xmark & \cmark & 64.7 & 88.6 \\

    \midrule
    ~~best combination & \cmark & \cmark & \xmark & \xmark & \cmark & \textbf{65.5} & \textbf{89.0} \\
    \midrule
    ~~All  & \cmark & \cmark & \cmark & \cmark & \cmark & 65.1 & 88.8 \\

    \bottomrule
\end{tabularx}

        \end{subtable}    

    &

        \begin{subtable}[t]{.5\linewidth}
            \vspace{-1 cm}
            \caption{\bf{Dataset-Task Agreement}}
            \centering
            \scriptsize
            \vspace{.25cm}  
            \includegraphics[width=.7\linewidth, height=.5\linewidth]{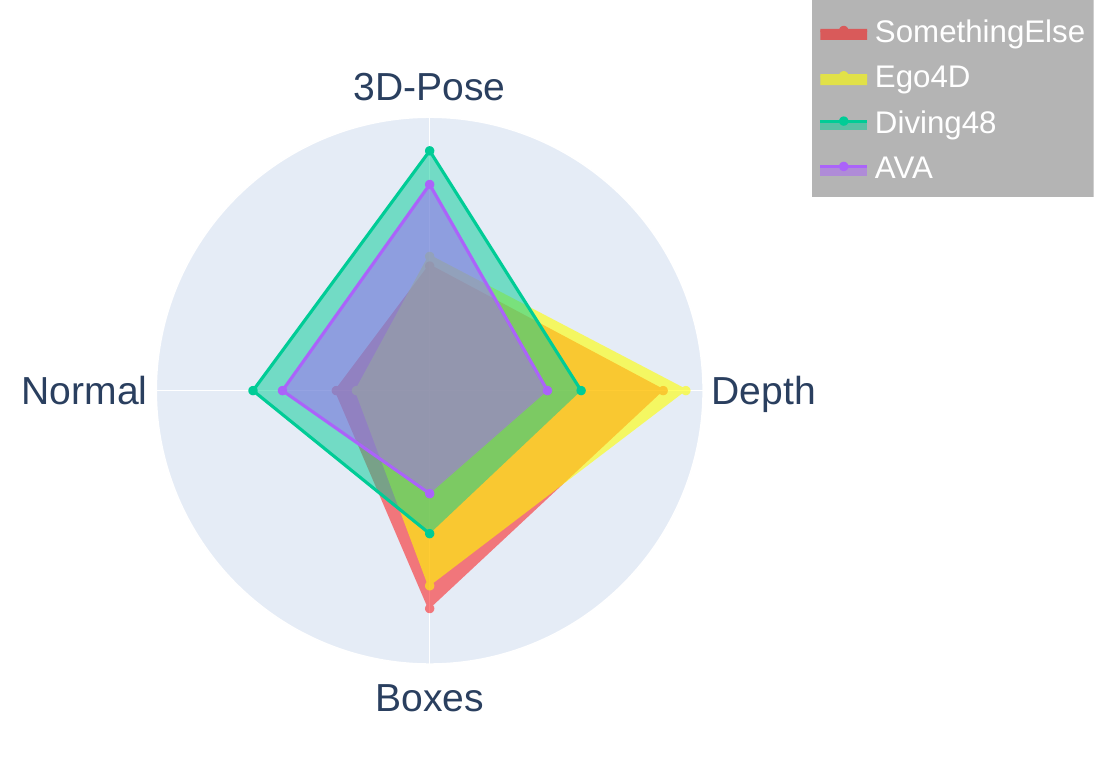}
            \label{fig:dataset-task}
        \end{subtable}

    \\

    \end{tabular}
    \caption{\textbf{Ablations.} We show (a) The Role of Prompts and Tuning. (b) Effect of Synthetic Data Size. (c) Contribution of Auxiliary Tasks. (d) Dataset-Task Agreement. A polygon represents a real video dataset, and the closer a vertex is to the circle border, the greater the gain from applying that synthetic task. The gains are scaled for comparison.}
    
    
    \label{tab:ablations}
    \vspace{-1.0em}
\end{table*}

\minisection{The Role of Prompts and Tuning} {\smodel} contains two main concepts: (i) the addition of multiple task-specific prompts dedicated to unique tasks. (ii) training these prompt representations to predict task-related labels from synthetic data. We present results for different combinations of these two factors in~\tabref{tab:abl_pvit}. First, to demonstrate the importance of having multiple prompts, one per task, we suggest the \emph{MViTv2 one-prompt (OP)} variant. This variant is similar to {\smodel} but uses \emph{a single prompt} instead of $n$ prompts for $n$ auxiliary tasks. Since the number of prompts decreases, we compensate by increasing the dimension size. As shown in \tabref{tab:abl_pvit}, PViT outperforms the OP variant, suggesting that multiple prompts are important for integrating information across tasks. 
\newline
Next, we consider the MViTv2 \emph{neutral-prompts (NP)} variant, which is simply MViTv2 with additional prompts but without additional synthetic supervision (similar to the MViTv2 VPT, but with an unfrozen backbone). The purpose of this variant is to examine whether the model performance is due to the increased model capacity. This result (63.4) is similar to the baseline without synthetic data (MViTv2, 63.3), suggesting that the gain of {\smodel} is due to the use of synthetic data. Last, the \emph{PViT VPT} variant is a simply {\smodel} with a frozen backbone. MViTv2 VPT differs from this variant since here, synthetic data is used for training. The result (53.9) emphasizes the importance of fine-tuning the backbone even when using synthetic data.

\minisection{Model Capacity and Efficiency Analysis} To determine whether the performance improvement is a result of increasing parameter size, \tabref{tab:abl_pvit} compares the number of parameters, FLOPS, and inference runtime between the methods. The main difference between the models is due to the additional task prompts and the task heads since the latter contains the most overhead (only during training). In our setting, task prompts only add ~20K parameters, while the task heads add 6.8M parameters. However, during test time, the heads are not used, and thus the parameter sizes are almost equal to the baseline (i.e., 38.2M), resulting in similar inference runtime and FLOPS as the baseline.

\minisection{Effect of Synthetic Data Size} Here, we examine the impact of the synthetic data portion on performance. In~\figgref{fig:ratios}, we plot the performance of {\smodel} as a function of the synthetic data portion when the largest value is obtained using all synthetic data. The positive slope suggests that adding synthetic data consistently improves results, which is an advantage since synthetic data is abundant. We note that the synthetic data we used is $\time3$ the size of the real data.

\minisection{Contribution from Auxiliary Tasks}
To investigate the impact of each auxiliary task on performance, we examined in \tabref{tab:abl_datatasks} how the auxiliary tasks contribute to performance individually, as well as the most effective combinations of auxiliary tasks. As can be seen, we find that performing {\smodel} on auxiliary tasks individually does improve 
performance (see also {\em Dataset Task Agreement} below). However, using all tasks (last line) improves more than any individual task, and is also close to the optimal combination. This reinforces our strategy of simply training on all tasks. 

\minisection{Dataset-Task Agreement} We next aim to explore how different synthetic tasks help real datasets. \figgref{fig:dataset-task} illustrates the gain for real datasets when trained on individual auxiliary tasks\footnote{The plot excludes segm. since it contributes equally to all datasets.}. It can be seen that the datasets are roughly clustered into two sets: (i) SomthingElse and Ego4D, which benefit more from Depth and Boxes. These datasets indeed contain hands interacting with \emph{objects} within close range of the camera and therefore having clearly expressed \emph{depth}. (ii) AVA and Diving48, which benefit more from Normals and Poses. These datasets generally consist of zoomed-out frames with mostly \emph{full human bodies} in scenes containing \emph{solid surfaces} (for example, pools, walls, etc.). For more details, see~\Secref{supp:expr:more_ablt} in the supplementary.

\minisection{Domain Gap Between Synthetic and Real Data}  In this work, we show that training {\smodel} on synthetic data leads to improved performance on real data. However, as synthetic and real data come from different domains, it is not apriori clear why the former should aid the latter. We hypothesize that our synthetic tasks are mostly low-level (e.g., depth/normal maps, segm. masks), and for these, there may be a smaller gap between synthetic and real domains (See~\cite{8953707,vu2019dada}). To illustrate this, we use our learned task heads to predict labels on real data. Recall that these heads are learned only on synthetic data. Figure~\ref{fig:vis_objs} shows results for this prediction, and it can be seen that the synthetic prompts predict well also on real data. This demonstrates that the synthetic tasks learned are also usable on real data. 



\section{Discussion and Limitations}

Semantic understanding of videos is a key element of human visual perception, but its modeling is still challenging for machine vision. In this work, we propose a new method for exploiting various types of scene-level data to improve the performance of video understanding tasks. We present a multi-task prompt learning approach for video transformers, where a shared transformer backbone is enhanced with task-specific prompts. The use of task-specific prompts allows the model to benefit from task-related information, among different domains. We demonstrate improved performance on several video understanding benchmarks, highlighting the effectiveness of the proposed approach. However, the multi-task prompt learning method is not necessarily limited to synthetic scene data, and thus we leave to future research the challenge of extending the work to train the method on real data as well as improving other downstream tasks in addition to video understanding. %

\vspace{-10pt}    
\subsubsection*{Acknowledgements}
\vspace{-5pt}
This project has received funding from the European Research Council (ERC) under the European Unions Horizon 2020 research and innovation programme (grant ERC HOLI 819080). Prof. Darrell’s group was supported in part by DoD including DARPA's Semafor, PTG and/or LwLL programs, as well as BAIR's industrial alliance programs. IBM research was supported by DARPA under Contract No. FA8750-19-C-1001.





{\small
\bibliographystyle{ieee_fullname}
\bibliography{egbib}

\begin{thebibliography}{100}\itemsep=-1pt

\bibitem{Armeni2017Joint2D}
Iro Armeni, Sasha Sax, Amir~Roshan Zamir, and Silvio Savarese.
\newblock Joint 2d-3d-semantic data for indoor scene understanding.
\newblock {\em ArXiv}, abs/1702.01105, 2017.

\bibitem{arnab2021vivit}
Anurag Arnab, Mostafa Dehghani, Georg Heigold, Chen Sun, Mario Lučić, and
  Cordelia Schmid.
\newblock Vivit: A video vision transformer, 2021.

\bibitem{arnab2021unified}
Anurag Arnab, Chen Sun, and Cordelia Schmid.
\newblock Unified graph structured models for video understanding.
\newblock In {\em ICCV}, 2021.

\bibitem{asai2022attempt}
Akari Asai, Mohammadreza Salehi, Matthew~E. Peters, and Hannaneh Hajishirzi.
\newblock Attentional mixtures of soft prompt tuning for parameter-efficient
  multi-task knowledge sharing.
\newblock {\em ArXiv}, abs/2205.11961, 2022.

\bibitem{avraham2022svit}
Elad~Ben Avraham, Roei Herzig, Karttikeya Mangalam, Amir Bar, Anna Rohrbach,
  Leonid Karlinsky, Trevor Darrell, and Amir Globerson.
\newblock Bringing image scene structure to video via frame-clip consistency of
  object tokens.
\newblock In {\em Thirty-Sixth Conference on Neural Information Processing
  Systems}, 2022.

\bibitem{Ba2016LayerN}
Jimmy Ba, Jamie~Ryan Kiros, and Geoffrey~E. Hinton.
\newblock Layer normalization.
\newblock {\em ArXiv}, abs/1607.06450, 2016.

\bibitem{2020ActionGraphs}
Amir Bar, Roei Herzig, Xiaolong Wang, Anna Rohrbach, Gal Chechik, Trevor
  Darrell, and A. Globerson.
\newblock Compositional video synthesis with action graphs.
\newblock In {\em ICML}, 2021.

\bibitem{baradel2018object}
Fabien Baradel, Natalia Neverova, Christian Wolf, Julien Mille, and Greg Mori.
\newblock Object level visual reasoning in videos.
\newblock In {\em ECCV}, pages 105--121, 2018.

\bibitem{battaglia2018relational}
Peter~W Battaglia, Jessica~B Hamrick, Victor Bapst, Alvaro Sanchez-Gonzalez,
  Vinicius Zambaldi, Mateusz Malinowski, Andrea Tacchetti, David Raposo, Adam
  Santoro, Ryan Faulkner, et~al.
\newblock Relational inductive biases, deep learning, and graph networks.
\newblock {\em arXiv preprint arXiv:1806.01261}, 2018.

\bibitem{gberta_2021_ICML}
Gedas Bertasius, Heng Wang, and Lorenzo Torresani.
\newblock Is space-time attention all you need for video understanding?
\newblock In {\em Proceedings of the International Conference on Machine
  Learning (ICML)}, July 2021.

\bibitem{Bhattacharjee2022MuITAE}
Deblina Bhattacharjee, Tong Zhang, Sabine S{\"u}sstrunk, and Mathieu Salzmann.
\newblock Muit: An end-to-end multitask learning transformer.
\newblock {\em 2022 IEEE/CVF Conference on Computer Vision and Pattern
  Recognition (CVPR)}, pages 12021--12031, 2022.

\bibitem{Brggemann2020AutomatedSF}
David Br{\"u}ggemann, Menelaos Kanakis, Stamatios Georgoulis, and Luc~Van Gool.
\newblock Automated search for resource-efficient branched multi-task networks.
\newblock {\em ArXiv}, abs/2008.10292, 2020.

\bibitem{detr2020}
Nicolas Carion, Francisco Massa, Gabriel Synnaeve, Nicolas Usunier, Alexander
  Kirillov, and Sergey Zagoruyko.
\newblock End-to-end object detection with transformers.
\newblock In Andrea Vedaldi, Horst Bischof, Thomas Brox, and Jan-Michael Frahm,
  editors, {\em Computer Vision -- ECCV 2020}, pages 213--229, 2020.

\bibitem{carreira2017quo}
Joao Carreira and Andrew Zisserman.
\newblock Quo vadis, action recognition? a new model and the kinetics dataset.
\newblock In {\em proceedings of the IEEE Conference on Computer Vision and
  Pattern Recognition}, pages 6299--6308, 2017.

\bibitem{Caruana1998MultitaskL}
Rich Caruana.
\newblock Multitask learning.
\newblock In {\em Encyclopedia of Machine Learning and Data Mining}, 1998.

\bibitem{CARUSO2017174}
L. Caruso, R. Russo, and S. Savino.
\newblock Microsoft kinect v2 vision system in a manufacturing application.
\newblock {\em Robotics and Computer-Integrated Manufacturing}, 48:174--181,
  2017.

\bibitem{8953707}
Yuhua Chen, Wen Li, Xiaoran Chen, and Luc Van~Gool.
\newblock Learning semantic segmentation from synthetic data: A geometrically
  guided input-output adaptation approach.
\newblock In {\em 2019 IEEE/CVF Conference on Computer Vision and Pattern
  Recognition (CVPR)}, pages 1841--1850, 2019.

\bibitem{Chen20214DContrastCL}
Yujin Chen, Matthias Nie{\ss}ner, and Angela Dai.
\newblock 4dcontrast: Contrastive learning with dynamic correspondences for 3d
  scene understanding.
\newblock {\em ArXiv}, abs/2112.02990, 2021.

\bibitem{Chen2020UNITERUI}
Yen-Chun Chen, Linjie Li, Licheng Yu, Ahmed~El Kholy, Faisal Ahmed, Zhe Gan, Yu
  Cheng, and Jingjing Liu.
\newblock Uniter: Universal image-text representation learning.
\newblock In {\em ECCV}, 2020.

\bibitem{Cheng2020PanopticDeepLabAS}
Bowen Cheng, Maxwell~D. Collins, Yukun Zhu, Ting Liu, Thomas~S. Huang, Hartwig
  Adam, and Liang-Chieh Chen.
\newblock Panoptic-deeplab: A simple, strong, and fast baseline for bottom-up
  panoptic segmentation.
\newblock {\em 2020 IEEE/CVF Conference on Computer Vision and Pattern
  Recognition (CVPR)}, pages 12472--12482, 2020.

\bibitem{Choutas2018PoTionPM}
Vasileios Choutas, Philippe Weinzaepfel, J{\'e}r{\^o}me Revaud, and Cordelia
  Schmid.
\newblock Potion: Pose motion representation for action recognition.
\newblock {\em 2018 IEEE/CVF Conference on Computer Vision and Pattern
  Recognition}, pages 7024--7033, 2018.

\bibitem{Crawshaw2020MultiTaskLW}
Michael Crawshaw.
\newblock Multi-task learning with deep neural networks: A survey.
\newblock {\em ArXiv}, abs/2009.09796, 2020.

\bibitem{DeSouza:Procedural:CVPR2017}
CR De~Souza, A Gaidon, Y Cabon, and AM Lopez~Pena.
\newblock Procedural generation of videos to train deep action recognition
  networks.
\newblock In {\em CVPR}, 2017.

\bibitem{Souza2017ProceduralGO}
C{\'e}sar~Roberto de Souza, Adrien Gaidon, Yohann Cabon, and Antonio
  Manuel~L{\'o}pez Pe{\~n}a.
\newblock Procedural generation of videos to train deep action recognition
  networks.
\newblock {\em 2017 IEEE Conference on Computer Vision and Pattern Recognition
  (CVPR)}, pages 2594--2604, 2017.

\bibitem{dosovitskiy2020vit}
Alexey Dosovitskiy, Lucas Beyer, Alexander Kolesnikov, Dirk Weissenborn,
  Xiaohua Zhai, Thomas Unterthiner, Mostafa Dehghani, Matthias Minderer, Georg
  Heigold, Sylvain Gelly, Jakob Uszkoreit, and Neil Houlsby.
\newblock An image is worth 16x16 words: Transformers for image recognition at
  scale.
\newblock {\em ICLR}, 2021.

\bibitem{du2015hierarchical}
Yong Du, Wei Wang, and Liang Wang.
\newblock Hierarchical recurrent neural network for skeleton based action
  recognition.
\newblock In {\em Proceedings of the IEEE conference on computer vision and
  pattern recognition}, pages 1110--1118, 2015.

\bibitem{mvit2021}
Haoqi Fan, Bo Xiong, Karttikeya Mangalam, Yanghao Li, Zhicheng Yan, Jitendra
  Malik, and Christoph Feichtenhofer.
\newblock Multiscale vision transformers.
\newblock In {\em ICCV}, 2021.

\bibitem{slowfast2019}
C. {Feichtenhofer}, H. {Fan}, J. {Malik}, and K. {He}.
\newblock Slowfast networks for video recognition.
\newblock In {\em 2019 IEEE/CVF International Conference on Computer Vision
  (ICCV)}, pages 6201--6210, 2019.

\bibitem{Gan2021ThreeDWorldAP}
Chuang Gan, Jeremy Schwartz, Seth Alter, Martin Schrimpf, James Traer,
  Julian~De Freitas, Jonas Kubilius, Abhishek Bhandwaldar, Nick Haber, Megumi
  Sano, Kuno Kim, Elias Wang, Damian Mrowca, Michael Lingelbach, Aidan Curtis,
  Kevin~T. Feigelis, Daniel Bear, Dan Gutfreund, David Cox, James~J. DiCarlo,
  Josh~H. McDermott, Joshua~B. Tenenbaum, and Daniel L.~K. Yamins.
\newblock Threedworld: A platform for interactive multi-modal physical
  simulation.
\newblock {\em ArXiv}, abs/2007.04954, 2021.

\bibitem{Gao2020DRGDR}
Chen Gao, Jiarui Xu, Yuliang Zou, and Jia-Bin Huang.
\newblock Drg: Dual relation graph for human-object interaction detection.
\newblock {\em ArXiv}, abs/2008.11714, 2020.

\bibitem{Gao2020MTLNASTN}
Yuan Gao, Haoping Bai, Zequn Jie, Jiayi Ma, Kui Jia, and Wei Liu.
\newblock Mtl-nas: Task-agnostic neural architecture search towards
  general-purpose multi-task learning.
\newblock {\em 2020 IEEE/CVF Conference on Computer Vision and Pattern
  Recognition (CVPR)}, pages 11540--11549, 2020.

\bibitem{Gao2019NDDRCNNLF}
Yuan Gao, Qi She, Jiayi Ma, Mingbo Zhao, W. Liu, and Alan~Loddon Yuille.
\newblock Nddr-cnn: Layerwise feature fusing in multi-task cnns by neural
  discriminative dimensionality reduction.
\newblock {\em 2019 IEEE/CVF Conference on Computer Vision and Pattern
  Recognition (CVPR)}, pages 3200--3209, 2019.

\bibitem{girdhar2019video}
Rohit Girdhar, Joao Carreira, Carl Doersch, and Andrew Zisserman.
\newblock Video action transformer network.
\newblock In {\em Proceedings of the IEEE Conference on Computer Vision and
  Pattern Recognition}, pages 244--253, 2019.

\bibitem{Girdhar2017ActionVLADLS}
Rohit Girdhar, Deva Ramanan, Abhinav~Kumar Gupta, Josef Sivic, and Bryan~C.
  Russell.
\newblock Actionvlad: Learning spatio-temporal aggregation for action
  classification.
\newblock {\em 2017 IEEE Conference on Computer Vision and Pattern Recognition
  (CVPR)}, pages 3165--3174, 2017.

\bibitem{goyal2017something}
Raghav Goyal, Samira~Ebrahimi Kahou, Vincent Michalski, Joanna Materzynska,
  Susanne Westphal, Heuna Kim, Valentin Haenel, Ingo Fruend, Peter Yianilos,
  Moritz Mueller-Freitag, et~al.
\newblock The" something something" video database for learning and evaluating
  visual common sense.
\newblock In {\em ICCV}, page~5, 2017.

\bibitem{Ego4D2021}
Kristen Grauman, Andrew Westbury, Eugene Byrne, Zachary Chavis, Antonino
  Furnari, Rohit Girdhar, Jackson Hamburger, Hao Jiang, Miao Liu, Xingyu Liu,
  Miguel Martin, Tushar Nagarajan, Ilija Radosavovic, Santhosh~Kumar
  Ramakrishnan, Fiona Ryan, Jayant Sharma, Michael Wray, Mengmeng Xu,
  Eric~Zhongcong Xu, Chen Zhao, Siddhant Bansal, Dhruv Batra, Vincent
  Cartillier, Sean Crane, Tien Do, Morrie Doulaty, Akshay Erapalli, Christoph
  Feichtenhofer, Adriano Fragomeni, Qichen Fu, Christian Fuegen, Abrham
  Gebreselasie, Cristina Gonzalez, James Hillis, Xuhua Huang, Yifei Huang,
  Wenqi Jia, Weslie Khoo, Jachym Kolar, Satwik Kottur, Anurag Kumar, Federico
  Landini, Chao Li, Yanghao Li, Zhenqiang Li, Karttikeya Mangalam, Raghava
  Modhugu, Jonathan Munro, Tullie Murrell, Takumi Nishiyasu, Will Price,
  Paola~Ruiz Puentes, Merey Ramazanova, Leda Sari, Kiran Somasundaram, Audrey
  Southerland, Yusuke Sugano, Ruijie Tao, Minh Vo, Yuchen Wang, Xindi Wu,
  Takuma Yagi, Yunyi Zhu, Pablo Arbelaez, David Crandall, Dima Damen,
  Giovanni~Maria Farinella, Bernard Ghanem, Vamsi~Krishna Ithapu, C.~V.
  Jawahar, Hanbyul Joo, Kris Kitani, Haizhou Li, Richard Newcombe, Aude Oliva,
  Hyun~Soo Park, James~M. Rehg, Yoichi Sato, Jianbo Shi, Mike~Zheng Shou,
  Antonio Torralba, Lorenzo Torresani, Mingfei Yan, and Jitendra Malik.
\newblock Ego4d: Around the {W}orld in 3,000 {H}ours of {E}gocentric {V}ideo.
\newblock {\em CoRR}, abs/2110.07058, 2021.

\bibitem{AVA2018}
Chunhui Gu, Chen Sun, David~A. Ross, Carl Vondrick, Caroline Pantofaru, Yeqing
  Li, Sudheendra Vijayanarasimhan, George Toderici, Susanna Ricco, Rahul
  Sukthankar, Cordelia Schmid, and Jitendra Malik.
\newblock {AVA:} {A} video dataset of spatio-temporally localized atomic visual
  actions.
\newblock In {\em 2018 {IEEE} Conference on Computer Vision and Pattern
  Recognition, {CVPR} 2018, Salt Lake City, UT, USA, June 18-22, 2018}, pages
  6047--6056. {IEEE} Computer Society, 2018.

\bibitem{resnet1}
Kaiming He, Xiangyu Zhang, Shaoqing Ren, and Jian Sun.
\newblock Deep residual learning for image recognition.
\newblock In {\em CVPR}, pages 770--778, 2016.

\bibitem{herzig2019canonical}
Roei Herzig, Amir Bar, Huijuan Xu, Gal Chechik, Trevor Darrell, and Amir
  Globerson.
\newblock Learning canonical representations for scene graph to image
  generation.
\newblock In {\em European Conference on Computer Vision}, 2020.

\bibitem{herzig2022orvit}
Roei Herzig, Elad Ben-Avraham, Karttikeya Mangalam, Amir Bar, Gal Chechik, Anna
  Rohrbach, Trevor Darrell, and Amir Globerson.
\newblock Object-region video transformers.
\newblock In {\em Conference on Computer Vision and Pattern Recognition
  (CVPR)}, 2022.

\bibitem{herzig2019stag}
Roei Herzig, Elad Levi, Huijuan Xu, Hang Gao, Eli Brosh, Xiaolong Wang, Amir
  Globerson, and Trevor Darrell.
\newblock Spatio-temporal action graph networks.
\newblock In {\em Proceedings of the IEEE International Conference on Computer
  Vision Workshops}, pages 0--0, 2019.

\bibitem{herzig2018mapping}
Roei Herzig, Moshiko Raboh, Gal Chechik, Jonathan Berant, and Amir Globerson.
\newblock Mapping images to scene graphs with permutation-invariant structured
  prediction.
\newblock In {\em Advances in Neural Information Processing Systems (NIPS)},
  2018.

\bibitem{Hinton2012dropout}
Geoffrey~E. Hinton, Nitish Srivastava, A. Krizhevsky, Ilya Sutskever, and R.
  Salakhutdinov.
\newblock Improving neural networks by preventing co-adaptation of feature
  detectors.
\newblock {\em ArXiv}, abs/1207.0580, 2012.

\bibitem{Hu2021UniTMM}
Ronghang Hu and Amanpreet Singh.
\newblock Unit: Multimodal multitask learning with a unified transformer.
\newblock {\em 2021 IEEE/CVF International Conference on Computer Vision
  (ICCV)}, pages 1419--1429, 2021.

\bibitem{hwang2020eldersim}
Hochul Hwang, Cheongjae Jang, Geonwoo Park, Junghyun Cho, and Ig-Jae Kim.
\newblock Eldersim: A synthetic data generation platform for human action
  recognition in eldercare applications, 2020.

\bibitem{Hwang2020ElderSimAS}
Hochul Hwang, Cheongjae Jang, Geonwoo Park, Junghyun Cho, and Ig-Jae Kim.
\newblock Eldersim: A synthetic data generation platform for human action
  recognition in eldercare applications.
\newblock {\em ArXiv}, abs/2010.14742, 2020.

\bibitem{Jerbi2020LearningOD}
Achiya Jerbi, Roei Herzig, Jonathan Berant, Gal Chechik, and Amir Globerson.
\newblock Learning object detection from captions via textual scene attributes.
\newblock {\em ArXiv}, abs/2009.14558, 2020.

\bibitem{ji2019action}
Jingwei Ji, Ranjay Krishna, Li Fei-Fei, and Juan~Carlos Niebles.
\newblock Action genome: Actions as composition of spatio-temporal scene
  graphs.
\newblock {\em arXiv preprint arXiv:1912.06992}, 2019.

\bibitem{ji2019largescale}
Yanli Ji, Feixiang Xu, Yang Yang, Fumin Shen, Heng~Tao Shen, and Wei-Shi Zheng.
\newblock A large-scale varying-view rgb-d action dataset for arbitrary-view
  human action recognition, 2019.

\bibitem{Jia2022VisualPT}
Menglin Jia, Luming Tang, Bor-Chun Chen, Claire Cardie, Serge Belongie, Bharath
  Hariharan, and Ser-Nam Lim.
\newblock Visual prompt tuning.
\newblock In {\em European Conference on Computer Vision (ECCV)}, 2022.

\bibitem{johnson2018image}
Justin Johnson, Agrim Gupta, and Li Fei-Fei.
\newblock Image generation from scene graphs.
\newblock In {\em Proceedings of the IEEE conference on computer vision and
  pattern recognition}, pages 1219--1228, 2018.

\bibitem{Kalogeiton2017ActionTD}
Vicky~S. Kalogeiton, Philippe Weinzaepfel, Vittorio Ferrari, and Cordelia
  Schmid.
\newblock Action tubelet detector for spatio-temporal action localization.
\newblock {\em 2017 IEEE International Conference on Computer Vision (ICCV)},
  pages 4415--4423, 2017.

\bibitem{kanazawa2018learning}
Angjoo Kanazawa, Jason~Y. Zhang, Panna Felsen, and Jitendra Malik.
\newblock Learning 3d human dynamics from video, 2018.

\bibitem{Kato2018CompositionalLF}
Keizo Kato, Yin Li, and Abhinav Gupta.
\newblock Compositional learning for human object interaction.
\newblock In {\em ECCV}, 2018.

\bibitem{kay2017kinetics}
Will Kay, Joao Carreira, Karen Simonyan, Brian Zhang, Chloe Hillier, Sudheendra
  Vijayanarasimhan, Fabio Viola, Tim Green, Trevor Back, Paul Natsev, et~al.
\newblock The kinetics human action video dataset.
\newblock {\em arXiv preprint arXiv:1705.06950}, 2017.

\bibitem{Kendall2018MultitaskLU}
Alex Kendall, Yarin Gal, and Roberto Cipolla.
\newblock Multi-task learning using uncertainty to weigh losses for scene
  geometry and semantics.
\newblock {\em 2018 IEEE/CVF Conference on Computer Vision and Pattern
  Recognition}, pages 7482--7491, 2018.

\bibitem{kim2020safcar}
Tae~Soo Kim and Gregory~D. Hager.
\newblock Safcar: Structured attention fusion for compositional action
  recognition, 2020.

\bibitem{kingma2014adam}
Diederik~P Kingma and Jimmy Ba.
\newblock Adam: A method for stochastic optimization.
\newblock {\em arXiv preprint arXiv:1412.6980}, 2014.

\bibitem{referential_relationships}
Ranjay Krishna, Ines Chami, Michael~S. Bernstein, and Li Fei{-}Fei.
\newblock Referring relationships.
\newblock {\em ECCV}, 2018.

\bibitem{leonardi2022egocentric}
Rosario Leonardi, Francesco Ragusa, Antonino Furnari, and Giovanni~Maria
  Farinella.
\newblock Egocentric human-object interaction detection exploiting synthetic
  data, 2022.

\bibitem{Lester2021ThePO}
Brian Lester, Rami Al-Rfou, and Noah Constant.
\newblock The power of scale for parameter-efficient prompt tuning.
\newblock In {\em Proceedings of the 2021 Conference on Empirical Methods in
  Natural Language Processing}, pages 3045--3059, Online and Punta Cana,
  Dominican Republic, Nov. 2021. Association for Computational Linguistics.

\bibitem{li2022uniformer}
Kunchang Li, Yali Wang, Peng Gao, Guanglu Song, Yu Liu, Hongsheng Li, and Yu
  Qiao.
\newblock Uniformer: Unified transformer for efficient spatiotemporal
  representation learning, 2022.

\bibitem{Li2019VisualBERTAS}
Liunian~Harold Li, Mark Yatskar, Da Yin, Cho-Jui Hsieh, and Kai-Wei Chang.
\newblock Visualbert: A simple and performant baseline for vision and language.
\newblock {\em ArXiv}, abs/1908.03557, 2019.

\bibitem{li2020oscar}
Xiujun Li, Xi Yin, Chunyuan Li, Xiaowei Hu, Pengchuan Zhang, Lei Zhang, Lijuan
  Wang, Houdong Hu, Li Dong, Furu Wei, Yejin Choi, and Jianfeng Gao.
\newblock Oscar: Object-semantics aligned pre-training for vision-language
  tasks.
\newblock {\em ECCV 2020}, 2020.

\bibitem{Li_2018_Diving48}
Yingwei Li, Yi Li, and Nuno Vasconcelos.
\newblock Resound: Towards action recognition without representation bias.
\newblock In {\em Proceedings of the European Conference on Computer Vision
  (ECCV)}, September 2018.

\bibitem{li2021improvedmvit}
Yanghao Li, Chao-Yuan Wu, Haoqi Fan, Karttikeya Mangalam, Bo Xiong, Jitendra
  Malik, and Christoph Feichtenhofer.
\newblock Mvitv2: Improved multiscale vision transformers for classification
  and detection.
\newblock In {\em CVPR}, 2022.

\bibitem{Liang2019PeekingIT}
Junwei Liang, Lu Jiang, Juan~Carlos Niebles, Alexander~G. Hauptmann, and Li
  Fei-Fei.
\newblock Peeking into the future: Predicting future person activities and
  locations in videos.
\newblock {\em 2019 IEEE/CVF Conference on Computer Vision and Pattern
  Recognition (CVPR)}, pages 5718--5727, 2019.

\bibitem{lin2018tsm}
Ji Lin, Chuang Gan, and Song Han.
\newblock Tsm: Temporal shift module for efficient video understanding, 2018.

\bibitem{kevin2022egovlp}
Kevin~Qinghong Lin, Alex~Jinpeng Wang, Mattia Soldan, Michael Wray, Rui Yan,
  Eric~Zhongcong Xu, Difei Gao, Rongcheng Tu, Wenzhe Zhao, Weijie Kong, et~al.
\newblock Egocentric video-language pretraining.
\newblock {\em arXiv preprint arXiv:2206.01670}, 2022.

\bibitem{Lin2019BMNBN}
Tianwei Lin, Xiao Liu, Xin Li, Errui Ding, and Shilei Wen.
\newblock Bmn: Boundary-matching network for temporal action proposal
  generation.
\newblock {\em 2019 IEEE/CVF International Conference on Computer Vision
  (ICCV)}, pages 3888--3897, 2019.

\bibitem{Liu2021ConflictAverseGD}
Bo Liu, Xingchao Liu, Xiaojie Jin, Peter Stone, and Qiang Liu.
\newblock Conflict-averse gradient descent for multi-task learning.
\newblock In {\em NeurIPS}, 2021.

\bibitem{Liu2021TowardsIM}
Liyang Liu, Yi Li, Zhanghui Kuang, Jing-Hao Xue, Yimin Chen, Wenming Yang,
  Qingmin Liao, and Wayne Zhang.
\newblock Towards impartial multi-task learning.
\newblock In {\em ICLR}, 2021.

\bibitem{liu2017enhanced}
Mengyuan Liu, Hong Liu, and Chen Chen.
\newblock Enhanced skeleton visualization for view invariant human action
  recognition.
\newblock {\em Pattern Recognition}, 68:346--362, 2017.

\bibitem{Liu2019EndToEndML}
Shikun Liu, Edward Johns, and Andrew~J. Davison.
\newblock End-to-end multi-task learning with attention.
\newblock {\em 2019 IEEE/CVF Conference on Computer Vision and Pattern
  Recognition (CVPR)}, pages 1871--1880, 2019.

\bibitem{liu2021videoswin}
Ze Liu, Jia Ning, Yue Cao, Yixuan Wei, Zheng Zhang, Stephen Lin, and Han Hu.
\newblock Video swin transformer.
\newblock {\em arXiv preprint arXiv:2106.13230}, 2021.

\bibitem{Loper2015SMPLAS}
Matthew Loper, Naureen Mahmood, Javier Romero, Gerard Pons-Moll, and Michael~J.
  Black.
\newblock Smpl: a skinned multi-person linear model.
\newblock {\em ACM Trans. Graph.}, 34:248:1--248:16, 2015.

\bibitem{materzynska2019something}
Joanna Materzynska, Tete Xiao, Roei Herzig, Huijuan Xu, Xiaolong Wang, and
  Trevor Darrell.
\newblock Something-else: Compositional action recognition with
  spatial-temporal interaction networks.
\newblock In {\em proceedings of the IEEE Conference on Computer Vision and
  Pattern Recognition}, 2020.

\bibitem{Mikami2021ASL}
Hiroaki Mikami, Kenji Fukumizu, Shogo Murai, Shuji Suzuki, Yuta Kikuchi, Taiji
  Suzuki, Shin ichi Maeda, and Kohei Hayashi.
\newblock A scaling law for synthetic-to-real transfer: How much is your
  pre-training effective?, 2021.

\bibitem{Mishra2022Task2SimTE}
Samarth Mishra, Rameswar Panda, Cheng~Perng Phoo, Chun-Fu Chen, Leonid
  Karlinsky, Kate Saenko, Venkatesh Saligrama, and Rog{\'e}rio~Schmidt Feris.
\newblock Task2sim: Towards effective pre-training and transfer from synthetic
  data.
\newblock {\em 2022 IEEE/CVF Conference on Computer Vision and Pattern
  Recognition (CVPR)}, pages 9184--9194, 2022.

\bibitem{Nagarajan2020EgoTopoEA}
Tushar Nagarajan, Yanghao Li, Christoph Feichtenhofer, and Kristen Grauman.
\newblock Ego-topo: Environment affordances from egocentric video.
\newblock {\em 2020 IEEE/CVF Conference on Computer Vision and Pattern
  Recognition (CVPR)}, pages 160--169, 2020.

\bibitem{patrick2021keeping}
Mandela Patrick, Dylan Campbell, Yuki~M. Asano, Ishan Misra~Florian Metze,
  Christoph Feichtenhofer, Andrea Vedaldi, and Joao~F. Henriques.
\newblock Keeping your eye on the ball: Trajectory attention in video
  transformers, 2021.

\bibitem{Peng2015LearningDO}
Xingchao Peng, Baochen Sun, Karim Ali, and Kate Saenko.
\newblock Learning deep object detectors from 3d models.
\newblock {\em 2015 IEEE International Conference on Computer Vision (ICCV)},
  pages 1278--1286, 2015.

\bibitem{Prakash2019StructuredDR}
Aayush Prakash, Shaad Boochoon, Mark Brophy, David Acuna, Eric Cameracci,
  Gavriel State, Omer Shapira, and Stan Birchfield.
\newblock Structured domain randomization: Bridging the reality gap by
  context-aware synthetic data.
\newblock {\em 2019 International Conference on Robotics and Automation
  (ICRA)}, pages 7249--7255, 2019.

\bibitem{Pumarola2021DNeRFNR}
Albert Pumarola, Enric Corona, Gerard Pons-Moll, and Francesc Moreno-Noguer.
\newblock D-nerf: Neural radiance fields for dynamic scenes.
\newblock {\em 2021 IEEE/CVF Conference on Computer Vision and Pattern
  Recognition (CVPR)}, pages 10313--10322, 2021.

\bibitem{Qiao2021ViPDeepLabLV}
Siyuan Qiao, Yukun Zhu, Hartwig Adam, Alan~Loddon Yuille, and Liang-Chieh Chen.
\newblock Vip-deeplab: Learning visual perception with depth-aware video
  panoptic segmentation.
\newblock {\em 2021 IEEE/CVF Conference on Computer Vision and Pattern
  Recognition (CVPR)}, pages 3996--4007, 2021.

\bibitem{raboh2020dsg}
Moshiko Raboh, Roei Herzig, Gal Chechik, Jonathan Berant, and Amir Globerson.
\newblock Differentiable scene graphs.
\newblock In {\em WACV}, 2020.

\bibitem{Rezatofighi_2018_CVPR}
Hamid Rezatofighi, Nathan Tsoi, JunYoung Gwak, Amir Sadeghian, Ian Reid, and
  Silvio Savarese.
\newblock Generalized intersection over union.
\newblock In {\em The IEEE Conference on Computer Vision and Pattern
  Recognition (CVPR)}, June 2019.

\bibitem{roberts:2021}
Mike Roberts, Jason Ramapuram, Anurag Ranjan, Atulit Kumar, Miguel~Angel
  Bautista, Nathan Paczan, Russ Webb, and Joshua~M. Susskind.
\newblock {Hypersim}: {A} photorealistic synthetic dataset for holistic indoor
  scene understanding.
\newblock In {\em International Conference on Computer Vision (ICCV) 2021},
  2021.

\bibitem{Ros2016TheSD}
Germ{\'a}n Ros, Laura Sellart, Joanna Materzynska, David V{\'a}zquez, and
  Antonio~M. L{\'o}pez.
\newblock The synthia dataset: A large collection of synthetic images for
  semantic segmentation of urban scenes.
\newblock {\em 2016 IEEE Conference on Computer Vision and Pattern Recognition
  (CVPR)}, pages 3234--3243, 2016.

\bibitem{sanh2022multitask}
Victor Sanh, Albert Webson, Colin Raffel, Stephen Bach, Lintang Sutawika, Zaid
  Alyafeai, Antoine Chaffin, Arnaud Stiegler, Arun Raja, Manan Dey, M~Saiful
  Bari, Canwen Xu, Urmish Thakker, Shanya~Sharma Sharma, Eliza Szczechla,
  Taewoon Kim, Gunjan Chhablani, Nihal Nayak, Debajyoti Datta, Jonathan Chang,
  Mike Tian-Jian Jiang, Han Wang, Matteo Manica, Sheng Shen, Zheng~Xin Yong,
  Harshit Pandey, Rachel Bawden, Thomas Wang, Trishala Neeraj, Jos Rozen,
  Abheesht Sharma, Andrea Santilli, Thibault Fevry, Jason~Alan Fries, Ryan
  Teehan, Teven~Le Scao, Stella Biderman, Leo Gao, Thomas Wolf, and Alexander~M
  Rush.
\newblock Multitask prompted training enables zero-shot task generalization.
\newblock In {\em International Conference on Learning Representations}, 2022.

\bibitem{relnets2017nips}
Adam Santoro, David Raposo, David~G Barrett, Mateusz Malinowski, Razvan
  Pascanu, Peter Battaglia, and Timothy Lillicrap.
\newblock A simple neural network module for relational reasoning.
\newblock In I. Guyon, U.~V. Luxburg, S. Bengio, H. Wallach, R. Fergus, S.
  Vishwanathan, and R. Garnett, editors, {\em Advances in Neural Information
  Processing Systems}, volume~30. Curran Associates, Inc., 2017.

\bibitem{Savva2019HabitatAP}
Manolis Savva, Abhishek Kadian, Oleksandr Maksymets, Yili Zhao, Erik Wijmans,
  Bhavana Jain, Julian Straub, Jia Liu, Vladlen Koltun, Jitendra Malik, Devi
  Parikh, and Dhruv Batra.
\newblock Habitat: A platform for embodied ai research.
\newblock {\em 2019 IEEE/CVF International Conference on Computer Vision
  (ICCV)}, pages 9338--9346, 2019.

\bibitem{shahroudy2016ntu}
Amir Shahroudy, Jun Liu, Tian-Tsong Ng, and Gang Wang.
\newblock Ntu rgb+d: A large scale dataset for 3d human activity analysis.
\newblock In {\em Proceedings of the IEEE conference on computer vision and
  pattern recognition}, pages 1010--1019, 2016.

\bibitem{sun2018actor}
Chen Sun, Abhinav Shrivastava, Carl Vondrick, Kevin Murphy, Rahul Sukthankar,
  and Cordelia Schmid.
\newblock Actor-centric relation network.
\newblock In {\em Proceedings of the European Conference on Computer Vision
  (ECCV)}, pages 318--334, 2018.

\bibitem{Sun2021TaskSN}
Guolei Sun, Thomas Probst, Danda~Pani Paudel, Nikola Popovic, Menelaos Kanakis,
  Jagruti~R. Patel, Dengxin Dai, and Luc~Van Gool.
\newblock Task switching network for multi-task learning.
\newblock {\em 2021 IEEE/CVF International Conference on Computer Vision
  (ICCV)}, pages 8271--8280, 2021.

\bibitem{Sun2020AdaShareLW}
Ximeng Sun, Rameswar Panda, and Rog{\'e}rio~Schmidt Feris.
\newblock Adashare: Learning what to share for efficient deep multi-task
  learning.
\newblock {\em ArXiv}, abs/1911.12423, 2020.

\bibitem{Sun2019VideoVR}
Xu Sun, Tongwei Ren, Yuan Zi, and Gangshan Wu.
\newblock Video visual relation detection via multi-modal feature fusion.
\newblock {\em Proceedings of the 27th ACM International Conference on
  Multimedia}, 2019.

\bibitem{Tan2019LXMERTLC}
Hao~Hao Tan and Mohit Bansal.
\newblock Lxmert: Learning cross-modality encoder representations from
  transformers.
\newblock In {\em EMNLP}, 2019.

\bibitem{tong2022videomae}
Zhan Tong, Yibing Song, Jue Wang, and Limin Wang.
\newblock Videomae: Masked autoencoders are data-efficient learners for
  self-supervised video pre-training, 2022.

\bibitem{touvron2021training}
Hugo Touvron, Matthieu Cord, Matthijs Douze, Francisco Massa, Alexandre
  Sablayrolles, and Herv{\'e} J{\'e}gou.
\newblock Training data-efficient image transformers \& distillation through
  attention.
\newblock In {\em International Conference on Machine Learning}, pages
  10347--10357. PMLR, 2021.

\bibitem{Vandenhende2020MTINetMT}
Simon Vandenhende, Stamatios Georgoulis, and Luc~Van Gool.
\newblock Mti-net: Multi-scale task interaction networks for multi-task
  learning.
\newblock In {\em ECCV}, 2020.

\bibitem{Varol2021SyntheticHF}
G{\"u}l Varol, Ivan Laptev, Cordelia Schmid, and Andrew Zisserman.
\newblock Synthetic humans for action recognition from unseen viewpoints.
\newblock {\em ArXiv}, abs/1912.04070, 2021.

\bibitem{varol21_surreact}
G{\"u}l Varol, Ivan Laptev, Cordelia Schmid, and Andrew Zisserman.
\newblock Synthetic humans for action recognition from unseen viewpoints.
\newblock In {\em IJCV}, 2021.

\bibitem{vu-etal-2022-spot}
Tu Vu, Brian Lester, Noah Constant, Rami Al-Rfou{'}, and Daniel Cer.
\newblock {SP}o{T}: Better frozen model adaptation through soft prompt
  transfer.
\newblock In {\em Proceedings of the 60th Annual Meeting of the Association for
  Computational Linguistics (Volume 1: Long Papers)}, pages 5039--5059, Dublin,
  Ireland, May 2022. Association for Computational Linguistics.

\bibitem{vu2019dada}
Tuan-Hung Vu, Himalaya Jain, Maxime Bucher, Mathieu Cord, and Patrick
  P{\'e}rez.
\newblock Dada: Depth-aware domain adaptation in semantic segmentation.
\newblock In {\em ICCV}, 2019.

\bibitem{Wang_videogcnECCV2018}
Xiaolong Wang and Abhinav Gupta.
\newblock Videos as space-time region graphs.
\newblock In {\em ECCV}, 2018.

\bibitem{Wang2020DifferentialTF}
Zhonghao Wang, Mo Yu, Yunchao Wei, Rog{\'e}rio~Schmidt Feris, Jinjun Xiong, Wen
  mei W.~Hwu, Thomas~S. Huang, and Humphrey Shi.
\newblock Differential treatment for stuff and things: A simple unsupervised
  domain adaptation method for semantic segmentation.
\newblock {\em 2020 IEEE/CVF Conference on Computer Vision and Pattern
  Recognition (CVPR)}, pages 12632--12641, 2020.

\bibitem{wang2022dualprompt}
Zifeng Wang, Zizhao Zhang, Sayna Ebrahimi, Ruoxi Sun, Han Zhang, Chen-Yu Lee,
  Xiaoqi Ren, Guolong Su, Vincent Perot, Jennifer Dy, et~al.
\newblock Dualprompt: Complementary prompting for rehearsal-free continual
  learning.
\newblock {\em European Conference on Computer Vision}, 2022.

\bibitem{wang2022learning}
Zifeng Wang, Zizhao Zhang, Chen-Yu Lee, Han Zhang, Ruoxi Sun, Xiaoqi Ren,
  Guolong Su, Vincent Perot, Jennifer Dy, and Tomas Pfister.
\newblock Learning to prompt for continual learning.
\newblock In {\em Proceedings of the IEEE/CVF Conference on Computer Vision and
  Pattern Recognition}, pages 139--149, 2022.

\bibitem{Kim2022HowTA}
Yo whan Kim, SouYoung Jin, Rameswar Panda, Hilde Kuehne, Leonid Karlinsky,
  Samarth Mishra, Venkatesh Saligrama, Kate Saenko, Aude Oliva, and
  Rog{\'e}rio~Schmidt Feris.
\newblock How transferable are video representations based on synthetic data?,
  2022.

\bibitem{lvu2021}
Chao-Yuan Wu and Philipp Kr\"{a}henb\"{u}hl.
\newblock {Towards Long-Form Video Understanding}.
\newblock In {\em {CVPR}}, 2021.

\bibitem{Wu2022MeMViTMM}
Chao-Yuan Wu, Yanghao Li, Karttikeya Mangalam, Haoqi Fan, Bo Xiong, Jitendra
  Malik, and Christoph Feichtenhofer.
\newblock Memvit: Memory-augmented multiscale vision transformer for efficient
  long-term video recognition.
\newblock {\em 2022 IEEE/CVF Conference on Computer Vision and Pattern
  Recognition (CVPR)}, pages 13577--13587, 2022.

\bibitem{Xia2018GibsonER}
F. Xia, Amir~Roshan Zamir, Zhi-Yang He, Alexander Sax, Jitendra Malik, and
  Silvio Savarese.
\newblock Gibson env: Real-world perception for embodied agents.
\newblock {\em 2018 IEEE/CVF Conference on Computer Vision and Pattern
  Recognition}, pages 9068--9079, 2018.

\bibitem{Xu2019LearningTD}
Bingjie Xu, Yongkang Wong, Junnan Li, Qi Zhao, and M. Kankanhalli.
\newblock Learning to detect human-object interactions with knowledge.
\newblock {\em 2019 IEEE/CVF Conference on Computer Vision and Pattern
  Recognition (CVPR)}, pages 2019--2028, 2019.

\bibitem{Xu2020reason}
Keyulu Xu, Jingling Li, Mozhi Zhang, Simon~S. Du, Ken ichi Kawarabayashi, and
  Stefanie Jegelka.
\newblock What can neural networks reason about?
\newblock In {\em International Conference on Learning Representations}, 2020.

\bibitem{Xu2022MultiTaskLW}
Yangyang Xu, Xiangtai Li, Haobo Yuan, Yibo Yang, Jing Zhang, Yunhai Tong, Lefei
  Zhang, and Dacheng Tao.
\newblock Multi-task learning with multi-query transformer for dense
  prediction.
\newblock {\em ArXiv}, abs/2205.14354, 2022.

\bibitem{Yu2020GradientSF}
Tianhe Yu, Saurabh Kumar, Abhishek Gupta, Sergey Levine, Karol Hausman, and
  Chelsea Finn.
\newblock Gradient surgery for multi-task learning.
\newblock {\em ArXiv}, abs/2001.06782, 2020.

\bibitem{zambaldi2018relational}
Vinicius Zambaldi, David Raposo, Adam Santoro, Victor Bapst, Yujia Li, Igor
  Babuschkin, Karl Tuyls, David Reichert, Timothy Lillicrap, Edward Lockhart,
  et~al.
\newblock Relational deep reinforcement learning.
\newblock {\em arXiv preprint arXiv:1806.01830}, 2018.

\bibitem{Zamir2018TaskonomyDT}
Amir~Roshan Zamir, Alexander Sax, Bokui~(William) Shen, Leonidas~J. Guibas,
  Jitendra Malik, and Silvio Savarese.
\newblock Taskonomy: Disentangling task transfer learning.
\newblock {\em 2018 IEEE/CVF Conference on Computer Vision and Pattern
  Recognition}, pages 3712--3722, 2018.

\bibitem{zhang2021tqn}
Chuhan Zhang, Ankush Gputa, and Andrew Zisserman.
\newblock Temporal query networks for fine-grained video understanding.
\newblock In {\em Conference on Computer Vision and Pattern Recognition
  (CVPR)}, 2021.

\bibitem{Structured_cvpr19}
Yubo Zhang, Pavel Tokmakov, Martial Hebert, and Cordelia Schmid.
\newblock A structured model for action detection.
\newblock In {\em 2019 IEEE/CVF Conference on Computer Vision and Pattern
  Recognition (CVPR)}, pages 9967--9976, 2019.

\bibitem{Zhao2022TubeRTT}
Jiaojiao Zhao, Yanyi Zhang, Xinyu Li, Hao Chen, Shuai Bing, Mingze Xu, Chunhui
  Liu, Kaustav Kundu, Yuanjun Xiong, Davide Modolo, Ivan Marsic, Cees G.~M.
  Snoek, and Joseph Tighe.
\newblock Tuber: Tubelet transformer for video action detection.
\newblock {\em 2022 IEEE/CVF Conference on Computer Vision and Pattern
  Recognition (CVPR)}, pages 13588--13597, 2022.

\bibitem{Zhou2022ConditionalPL}
Kaiyang Zhou, Jingkang Yang, Chen~Change Loy, and Ziwei Liu.
\newblock Conditional prompt learning for vision-language models.
\newblock {\em 2022 IEEE/CVF Conference on Computer Vision and Pattern
  Recognition (CVPR)}, pages 16795--16804, 2022.

\bibitem{Zhou2022LearningTP}
Kaiyang Zhou, Jingkang Yang, Chen~Change Loy, and Ziwei Liu.
\newblock Learning to prompt for vision-language models.
\newblock {\em Int. J. Comput. Vis.}, 130:2337--2348, 2022.

\bibitem{Zhou2020PatternStructureDF}
Lingli Zhou, Zhen Cui, Chunyan Xu, Zhenyu Zhang, Chaoqun Wang, Tong Zhang, and
  Jian Yang.
\newblock Pattern-structure diffusion for multi-task learning.
\newblock {\em 2020 IEEE/CVF Conference on Computer Vision and Pattern
  Recognition (CVPR)}, pages 4513--4522, 2020.

\end{thebibliography}
}


\newpage
\appendix
\clearpage
\renewcommand{\thefootnote}{\fnsymbol{footnote}}
\setcounter{page}{1} 
\setcounter{section}{0} 


\twocolumn[ 
\Large\textbf{Supplementary Material for PViT} 
\maketitle 
] 


\section*{}
In this supplementary file, we provide additional information about our experimental results, qualitative examples, implementation details and datasets. Specifically, \Secref{supp:expr} provides more experiment results, \Secref{supp:qual} provides qualitative visualizations to illustrate our approach, \Secref{supp:impl} provides additional implementation details, and \Secref{supp:datasets} provides additional datasets details.


\section{Additional Experiment Results}
\label{supp:expr}

We begin by presenting additional baseline results for all datasets and tasks in (\Secref{supp:expr:baselines}). Next, we present additional ablations (\Secref{supp:expr:more_ablt}) we performed in order to test the contribution of the different {\smodel} components.

\subsection{Baselines Comparison}
\label{supp:expr:baselines}

{Here, we evaluate several alternative ViT approaches (\emph{MViTv2 MT} and \emph{MViTv2 VPT}) to our task of using synthetic data towards improving action recognition models. Additionally, we report additional baselines that are comparable in compute and size to further compare to other approaches in (see~\tabref{supp:table:baselines}), such as ORViT Mformer~\cite{herzig2022orvit}, UniFormer-S~\cite{li2022uniformer}, SViT~\cite{avraham2022svit}, VideoMAE~\cite{tong2022videomae}, Video SWIN Transformer~\cite{liu2021videoswin}, STIN~\cite{materzynska2019something}, and SAFCAR~\cite{kim2020safcar}. We can observe that our PViT approach improves upon MViTv2 and is competitive with other strong models. We note that even compared to VideoMAE, a recent self-supervised learning method, our results are similar in AVA (+1.3) and SSv2 (-0.3), although VideoMAE utilizes a larger backbone and more computing for training. Finally, PViT can be applied to any pretrained backbone, which gives it an advantage over other methods.
}



\subsection{Additional Ablations}
\label{supp:expr:more_ablt}

Next, we provide additional ablations that further illustrates the benefits of our {\smodel}.

\minisection{The importance of synthetic scene data} To examine how important the information provided by the synthetic scene data is, we test the {\smodel} model, but provide it with ``useless'' synthetic label information. Specifically, we run an experiment in which the synthetic scene annotations are shuffled. As a result, the ground truth of the instance-level is shuffled for each synthetic scene task (e.g., for dense prediction tasks, the GT maps are shuffled). This ablation obtained 63.4\%, similar to the baseline (63.3\%). This is expected since wrong scene annotations are not likely to provide additional benefit beyond the baseline. Moreover, the model is capable of ignoring prompts if they are not required, so they should not have a negative impact beyond the baseline.




\minisection{Prompts for real-world data} Even though real-world datasets are less reach in annotations compared to synthetic, {\smodel} can still use them if available. To examine this, we added 2D hand-object boxes from SomethingElse as an additional auxiliary task along with its own prompt. This improved results by +1.2, suggesting that real data, if available, is beneficial. Clearly, the combination of synthetic and real data offers many promising and interesting directions, and we leave those to future work.


\minisection{Comparison to a pretraining approach} 
Another approach for using synthetic datasets is first to pretrain on the synthetic data, and then finetune on the video-related task. Here, we demonstrate the effectiveness of our PViT approach as compared to this standard pretraining approach. To implement pretraining, we add prediction heads on top of MViTv2, and train them only on the synthetic datasets. Next, we remove these prediction heads and finetune the model by predicting using the CLS token. This approach achieved 61.9\% compared to 63.3\% for MViTv2 baseline and 65.5\% for our {\smodel} approach. This indicates that our {\smodel} approach utilizes task information more effectively than a standard pretraining approach.


\minisection{Number of task prompts} This ablation tests whether adding more prompts per task will improve the results compared to {\smodel}, which uses one prompt per task. We add a total of 20 prompts to each task, which results in 65.4\%, demonstrating that the addition of more prompts does not necessarily improve its performance. Clearly, there are many possible design choices, such as selecting a number of prompts per task, their dimension, integrating into different depths, etc., and we leave those to future work.

\begin{figure}[t!]
    \centering
    \includegraphics[width=1.0\linewidth]{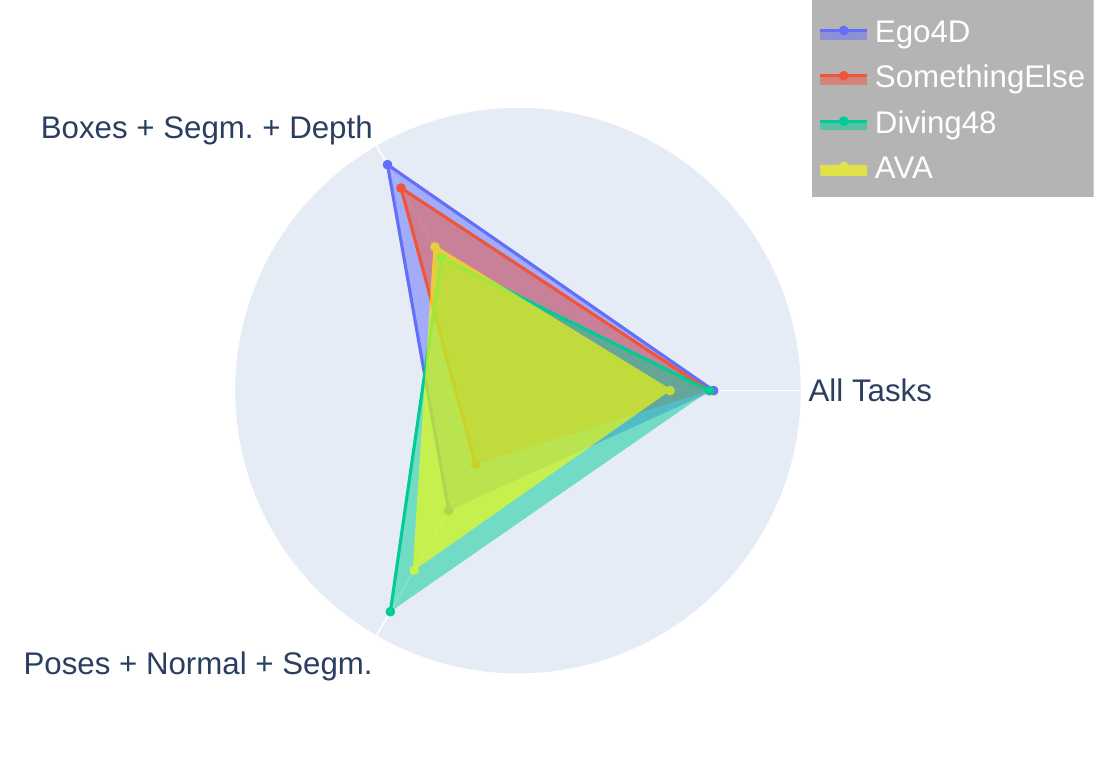}
    \captionof{figure}{Dataset-Task Agreement. A polygon represents a real video dataset, and the closer a vertex is to the circle border, the greater the gain from using that synthetic task. The gains are scaled for comparison.}
    \label{supp:fig:dataset_task_agreement}
\end{figure}

\renewcommand{\thefootnote}{\fnsymbol{footnote}}
\begin{table*}[tb!]
    \centering
    \begin{subtable}[t]{.34\linewidth}
	\tablestyle{3pt}{1.25}
	\scriptsize
	    \centering
        \caption{\bf{Something--Something V2}}
	    \label{supp:tab:sota_ssv2}
		\begin{tabularx}
{\linewidth}{@{}l c c c}
    \toprule
    {Model} & Pretrain & {Top-1} & {Top-5}\\ 
    \midrule
    SlowFast~\cite{slowfast2019}, R101 & K400 & 63.1 & 87.6 \\
    {MViTv1~\cite{mvit2021}} & {\scriptsize K400}  &  64.7 &  89.2 \\
    ViViT-L~\cite{arnab2021vivit} & {\scriptsize IN+K400}  & 65.4 & 89.8 \\
    {UniFormer-S~\cite{li2022uniformer}} & {\scriptsize IN+K600}  & 67.9 & 92.1 \\
    {ORViT Mformer~\cite{herzig2022orvit}} & {\scriptsize K400}  & 67.9 & 90.5 \\
    {VideoMAE (ViT-S)} & {\scriptsize K400}  &  66.8 &  90.3 \\
    {MViTv2~\cite{li2021improvedmvit}} & {\scriptsize K400}  &  68.2 &  91.4 \\
    \midrule

    
    
    {MViTv2 MT} & {\scriptsize K400}  &  68.4 &  91.3 \\
    {MViTv2 VPT} & {\scriptsize K400}  &  61.5 &  87.5 \\
    \midrule

    \bf PViT (Ours) & {\scriptsize K400}  & \textbf{69.6}\gcol{$+$1.2}  & \textbf{91.6}\gcol{$+$0.2} \\

    \bottomrule
\end{tabularx}
	\end{subtable}
    \begin{subtable}[t]{.30\linewidth}
    \tablestyle{3pt}{1.39}
    \scriptsize
        \centering
		\caption{\bf{Diving48}}
        \label{supp:tab:sota_diving48}
		\renewcommand*{\thefootnote}{\fnsymbol{footnote}}
\setlength{\tabcolsep}{3pt}

\begin{tabularx}
{\linewidth}{@{}l c c c}
    \toprule
    {Model} & {Pretrain~} & {Frames} & {Top-1} \\ 
    \midrule

    SlowFast~\cite{slowfast2019}, R101 & K400 & 16 & 77.6 \\
    TimeSformer~\cite{gberta_2021_ICML} & IN & 16 & 74.9 \\
    TimeSformer-L~\cite{gberta_2021_ICML} & IN & 96 & 81.0 \\
    SViT~\cite{avraham2022svit} & K400 & 16 & 79.8\\
    MViTv2~\cite{li2021improvedmvit} & K400 & 16 & 73.1\\
    
    \midrule
    MViTv2 MT & K400 & 16 & 75.6 \\
    MViTv2 VPT & K400 & 16 & 69.8 \\
    \midrule
    \textbf{PViT (Ours)} & K400 & 16 & \textbf{85.8}\gcol{+6.0} \\

    \bottomrule
\end{tabularx}

	\end{subtable}
	\begin{subtable}[t]{.30\linewidth}
	\tablestyle{6pt}{1.25}
	\scriptsize
	    \centering
	    \caption{\bf{AVA-V2.2}}
	    \label{supp:tab:sota_ava}
	\begin{tabular}{l c c c}
	    \toprule
		{Model} & {Pretrain} & {mAP} \\ 
		\midrule
		{SlowFast~\cite{slowfast2019}}, R50 & {K400} & 22.7 \\ 
		{SlowFast~\cite{slowfast2019}}, R101 & {K400} & 23.8 \\ 
		{ORViT MViT-B~\cite{herzig2022orvit}}  & {K400} & 26.6 \\ 
		{VideoMAE (ViT-S)~\cite{tong2022videomae}}  & {K400} & 22.5 \\ 
		{VideoMAE (ViT-B)~\cite{tong2022videomae}}  & {K400} & 26.7 \\ 
		{MViTv1~\cite{mvit2021}}  & {K400} & 25.5 \\ 
		{MViTv2~\cite{li2021improvedmvit}}  & {K400} & 26.8 \\
        \midrule
        {MViTv2 MT} & {K400} & 27.2 \\
        {MViTv2 VPT} & {K400} & 19.0 \\
		\midrule
		\textbf{PViT (Ours)} & {K400} & \textbf{28.4}\gcol{+1.6} \\
		\bottomrule
	\end{tabular}

	\end{subtable}
 
	\begin{subtable}[t]{.45\linewidth}
	\tablestyle{3pt}{1.25}
	\scriptsize
	    \centering
	    \caption{\bf{SomethingElse}}
	    \label{supp:tab:smthelse}
	    \begin{tabular}{l cc cc cc cc}
        \toprule
        
        \multirow{2}{*}{{Model}} & \multicolumn{2}{c}{Compositional} & \multicolumn{2}{c}{Base} & \multicolumn{2}{c}{Few-Shot} \\ 
        
         & {Top-1} & {Top-5} & {Top-1} & {Top-5} & {5-Shot} & {10-Shot}  \\
        
        \midrule
        {I3D~\cite{carreira2017quo}} & 42.8 & 71.3 & 73.6 & 92.2 & 21.8 & 26.7 \\
        {SlowFast~\cite{slowfast2019}} & 45.2 & 73.4 & 76.1 & 93.4 & 22.4 & 29.2 \\
        {TimeSformer~\cite{gberta_2021_ICML}} & 44.2 & 76.8 & 79.5 & 95.6 & 24.6 & 33.8 \\
        {STIN~\cite{materzynska2019something}} & 48.2 & 72.6 & - & - & - & - \\
        {TSM ~\cite{lin2018tsm}} & 52.3 & 78.0 & - & - & - & - \\
        {Mformer~\cite{patrick2021keeping}} &  60.2 & 85.8 & 82.8 & 96.2 & 28.9 & 33.8 \\
        {SAFCAR~\cite{kim2020safcar}} & 60.7 & 84.2 & - & - & - & - \\
        {MViTv2~\cite{li2021improvedmvit}} &  63.3 & 87.5 & 83.7 & 96.8 & 32.7 & 40.2 \\
        
        \midrule
        MViTv2 MT & 62.7 & 87.6 & 81.4 & 96.2 & 34.0 & 40.9 \\
        MViTv2 VPT & 53.0 & 81.8 & 76.8 & 94.8 & 31.8 & 39.0 \\
        \midrule
        \multirow{2}{*}{\textbf{PViT (Ours)}} & \textbf{65.5} & \textbf{89.0} & \textbf{85.0} & \textbf{97.4} & \textbf{34.3} & \textbf{41.3} \\
        & \gcol{$+$2.2} & \gcol{$+$2.5} & \gcol{$+$1.3} & \gcol{$+$0.6} & \gcol{$+$1.6} & \gcol{$+$1.1} \\
        \bottomrule
\end{tabular}%


    


	\end{subtable}
	\begin{subtable}[t]{.35\linewidth}
	\tablestyle{3pt}{1.25}
	\scriptsize
	    \centering
	    \caption{\bf{Ego4D}}
	    \label{supp:tab:ego4d}
		


	\begin{tabular}{l c c c c}
	    \toprule
		\multirow{2}*{Model} & {Temporal} & {PNR}
		\\ & {Localization Error} & {Classification Top-1}
		\\
		\midrule
		
        {Bi-LSTM}           & 0.790 & 65.3 \\
        {BMN~\cite{Lin2019BMNBN}}               & 0.780 & - \\
        {I3D ResNet-50~\cite{carreira2017quo}}     & 0.739 & 68.7 \\
        {EgoVLP (TimeSformer)~\cite{kevin2022egovlp}}           & 0.666 & 73.9 \\
        {Video Swin Transformer~\cite{liu2021videoswin}}           & 0.660 & 69.5 \\
        {MViTv2~\cite{li2021improvedmvit}}           & 0.702 & 71.6 \\
        \midrule
        {MViTv2 MT}           & 0.640 & 73.6 \\
        {MViTv2 OP}           & 0.652 & 73.7 \\
        \midrule
		\textbf{PViT (Ours)} & 0.637\gcol{-0.065} & \textbf{74.8}\gcol{+3.2} \\
		\bottomrule
	\end{tabular}

	\end{subtable}

    \caption{\textbf{Results on SSv2, Diving48, AVA-V2.2, SomethingElse, and Ego4D datasets.} We report top-1 and top-5 accuracy on SSv2 and SomethingElse. On AVA, we report the mAP metric. On Diving48, we report top-1. On Ego4D we report classification error. IN refers to ImageNet-21K.} 
    \label{supp:table:baselines}
\end{table*}
\renewcommand{\thefootnote}{\arabic{footnote}}



\minisection{Dataset-Task Agreement} In~\figgref{supp:fig:dataset_task_agreement}, we aim to explore how a different synthetic task combination helps real datasets. Since there are multiple possible subsets, we simplify and focus on only two subsets: $S_1$ = \{Boxes, Segmentation, Depth\} and $S_2$ = \{Poses, Normal, Segmentation\}. The former relates to hand-object interaction (HOI), and the latter to human action (HA). The figure shows the accuracy for real datasets when trained on either $S_1$, $S_2$, or all five tasks. This confirms our original hypothesis from the main paper that the datasets are roughly clustered into two categories: (i) SomthingElse and Ego4D benefit more from the HOI set. These datasets indeed usually contain hands interacting with objects, often in first person and with a low field of view. (ii) AVA and Diving48 belong benefit more from the HA group. These datasets generally consist of zoomed-out frames with mostly full human bodies.

\minisection{Contribution from Datasets and Tasks} In order to quantify the impact of each dataset and task, we conducted a comprehensive analysis in \tabref{supp:tab:res_per_dataset}. At the top of the table we display the contribution of each synthetic dataset to the downstream task, and at the bottom we display the contribution of each synthetic task (namely, we use all existing annotations from across all of our synthetic datasets). We observe that EHOI achieves the highest gains. This is similar to our observation in the main paper that hand-object interaction videos (HOI) benefit more from bounding box supervision. For more details, see the Dataset-Task Agreement ablation (\figgref{fig:dataset-task}) in the main paper. 
\newline
In the bottom portion of the table, we examined in the auxiliary tasks contribute to performance individually, as well as the most effective combinations of auxiliary tasks. As can be seen, we find that performing {\smodel} on auxiliary tasks individually does improve performance (see also {\em Dataset Task Agreement} below). However, using all tasks (last line) improves more than any individual task, and is also close to the optimal combination. This reinforces our strategy of simply training on all tasks. For a visualization of the datasets, see~\Secref{supp:qual} in supplementary.


\begin{table}[t!]

    \centering
	\tablestyle{1.0pt}{1.0}

	\begin{tabular}{@{}l c c c c c c c c r@{} r}

        \toprule
        ~~\multirow{2}{*}{Dataset} & ~~\multirow{2}{*}{Depth} & ~~\multirow{2}{*}{Segm.} & ~~\multirow{2}{*}{Normal} & {3D} & {2D} & ~~\multirow{2}{*}{Top-1} & ~~\multirow{2}{*}{Top-5} \\
         & & & & {Poses} & {Boxes} & \\
        \midrule
        ~~-      & \xmark & \xmark & \xmark & \xmark & \xmark & 63.3 & 87.5 \\
        \midrule
        ~~PHAV      & \cmark & \cmark & \xmark & \xmark & \xmark & 64.2 & 87.6 \\
        ~~SUR  & \cmark & \xmark & \xmark & \cmark & \xmark & 63.9 & 88.4 \\
        ~~ES & \xmark & \xmark & \xmark & \cmark & \xmark & 63.9 & 88.1 \\
        ~~HS  & \cmark & \xmark & \cmark & \xmark & \xmark & 64.1 & 87.4 \\
        ~~EHOI      & \xmark & \cmark & \xmark & \xmark & \cmark & 65.0 & 88.5 \\
        \midrule
        ~~PHAV+HS+SURR  & \cmark & \xmark & \xmark & \xmark & \xmark & 64.8 & 88.7 \\
        ~~SUR+EHOI  & \xmark & \cmark & \xmark & \xmark & \xmark & 65.0 & 88.7
        \\
        ~~HS  & \xmark & \xmark & \cmark & \xmark & \xmark & 63.9 & 88.2 \\
        ~~SUR+ES  & \xmark & \xmark & \xmark & \cmark & \xmark & 64.1 & 88.4 \\
        ~~EHOI      & \xmark & \xmark & \xmark & \xmark & \cmark & 64.7 & 88.6 \\

        \midrule
        ~~best combination & \cmark & \cmark & \xmark & \xmark & \cmark & \textbf{65.5} & \textbf{89.0} \\    
        
        \midrule
        ~~All  & \cmark & \cmark & \cmark & \cmark & \cmark & 65.1 & 88.8 \\
    
        \bottomrule
    \end{tabular}
    
	\caption{\textbf{Compositional Action Recognition task on the SomethingElse dataset.} The contribution of every synthetic auxiliary dataset (top) and task (bottom).} 
	\label{supp:tab:res_per_dataset}
\end{table}

\section{Qualitative Visualizations}
\label{supp:qual}

\begin{figure*}[t!]
    \centering
    \includegraphics[width=0.85\linewidth]{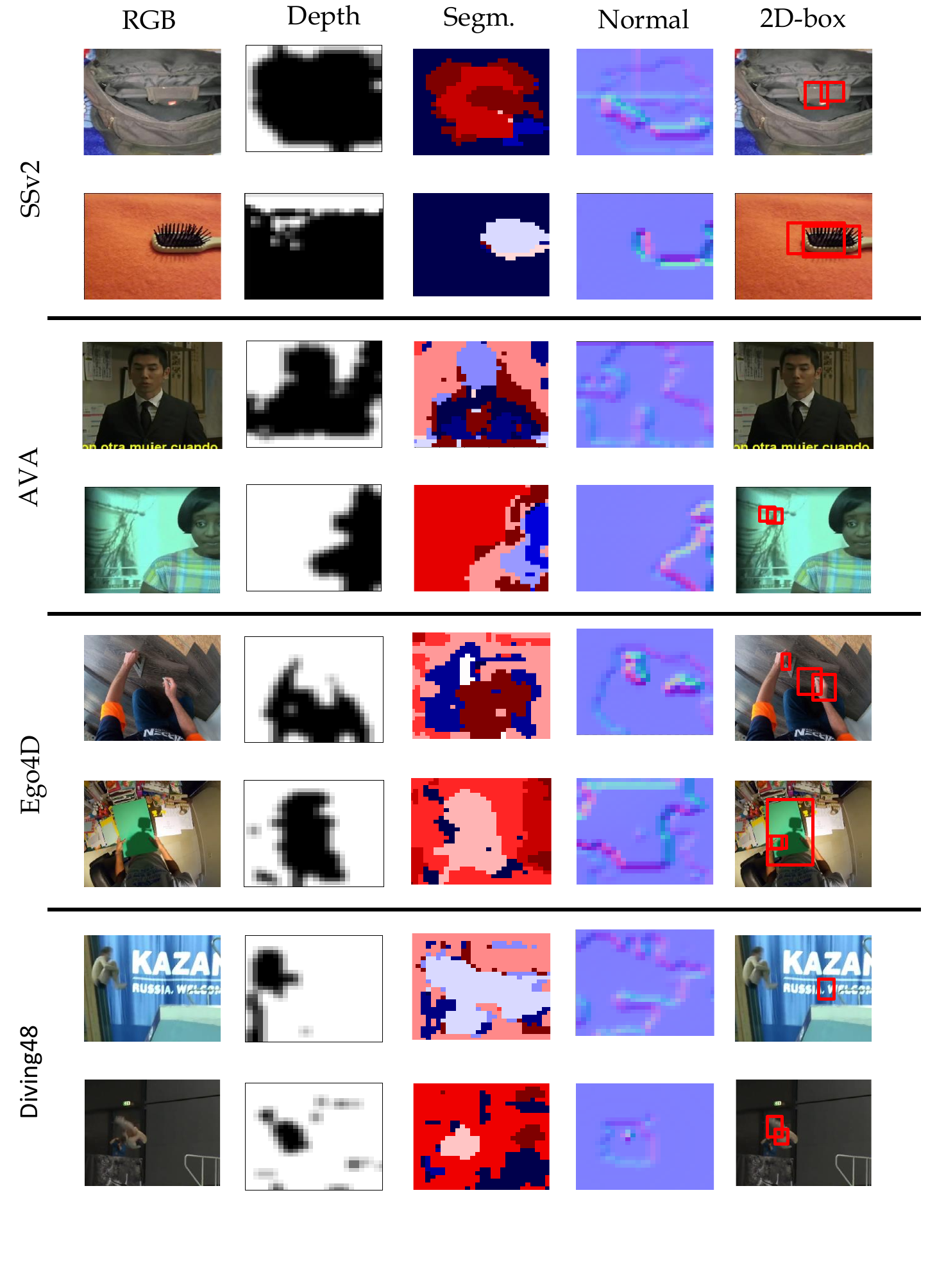}
    \vspace{-1.6cm}
    \captionof{figure}{\textbf{Qualitative visualization of the ``Task Prompts''}. Visualization of the output of the ``task prompts'' prediction heads on frames from the SSv2, Diving48, Ego4D, and AVA datasets. The model was trained on the SomethingElse dataset for action recognition. The predictions are the head outputs, $H_i$, for depth, normal, part-semantic segmentation and hand-object 2D boxes. It can be observed that the task prompts produce meaningful maps, despite not receiving labels for the real videos.} 
    \label{supp:fig:more_vis}
\end{figure*}

\figgref{supp:fig:more_vis} and~\figgref{fig:vis_objs} in the main paper show  visualizations of ``task prompts'' predictions on examples of real videos from SSv2, Diving48, Ego4D, and AVA. It can be seen that predictions are reasonable, despite the model not being trained on these labels for the real videos. For better illustration, we show in \figgref{supp:fig:data_vis} the different auxiliary synthetic datasets we used in the main paper, as described in~\Secref{sec:expr:datasets} and further elaborated upon in~\Secref{supp:datasets:aux}.

\section{Additional Implementation Details}
\label{supp:impl}

Our {\smodel} model can be used on top of the most common video transformers (MViT~\cite{mvit2021}, TimeSformer~\cite{gberta_2021_ICML}, Mformer~\cite{patrick2021keeping}, Video Swin~\cite{liu2021videoswin}). For our experiments, we choose the MViTv2~\cite{li2021improvedmvit} model because it performs well empirically. These are all implemented based on the MViTv2~\cite{li2021improvedmvit} library (available at \url{https://github.com/facebookresearch/mvit}), and we implement {\smodel} based on this repository. Furthermore, we set the $\lambda$ parameters (see Equation \ref{sec:eq:all_losses}) for the $\mathcal{L}_{Depth}$, $\mathcal{L}_{Normal}$, $\mathcal{L}_{Segm}$, $\mathcal{L}_{3DPose}$, $\mathcal{L}_{Boxes}$, and $\mathcal{L}_{DT}$ losses, to $0.5$, $0.5$, $0.1$, $3.0$, $0.1$ and $1$ respectively (across all datasets). We choose these lambda components such that all loss components have the same scale. We elaborate next on the additional implementation details for each dataset, including information about optimization, and training and inference.

\minisection{Dense Prediction Heads} In order to preserve the spatio-temporal information in dense prediction tasks, we use patch tokens in addition to task the tokens. First, we upsample patch tokens from layers 2, 12, 15 (out of 16) using a 3D convolution layer, followed by Dropout and concatenation. We then concatenate them with relevant task tokens and forward them to an MLP for a final prediction.

\subsection{Diving48}
\label{supp:impl:diving48}

\minisection{Dataset} Diving48~\cite{Li_2018_Diving48} contains 16K training and 3K testing videos spanning 48 fine-grained diving categories of diving activities. For all of these datasets, we use standard classification accuracy as our main performance metric.

\minisection{Optimization details} We train using $16$ frames with sample rate $4$ and batch-size $128$ (comprising $64$ videos and $64$ auxiliary synthetic datasets) on $8$ RTX 3090 GPUs. We train our network for 10 epochs with Adam optimizer~\cite{kingma2014adam} with a momentum of $9e-1$ and Gamma $1e-1$. Following~\cite{li2021improvedmvit}, we use $lr = 1.5e-4$ with half-period cosine decay.

\minisection{Training details} We use crops of size $224$ for the standard model and jitter scales between $256-320$. together with RandomFlip augmentation. Finally, we sample $T$ frames from the start and end annotation times, following ~\cite{zhang2021tqn}. 

\minisection{Inference details} We take 3 spatial crops per single clip to form predictions over a single video in testing, as in~\cite{gberta_2021_ICML}.

\subsection{SomethingElse}
\label{supp:impl:smthelse}
\minisection{Dataset} The SomethingElse dataset~\cite{materzynska2019something} contains 174 action categories with 54,919 training and 57,876 validation samples. The compositional~\cite{materzynska2019something} split in this dataset provides disjoint combinations of a verb (action) and noun (object) in the training and testing set, defining two disjoint groups of nouns $\{\mathcal{A}, \mathcal{B}\}$ and verbs $\{1, 2\}$. Given the splits of groups, they combine the training set as $1\mathcal{A} + 2\mathcal{B}$, while the validation set is constructed by flipping the combination into $1\mathcal{B} + 2\mathcal{A}$. In this way, different combinations of verbs and nouns are divided into training or testing splits.

\minisection{Few Shot Compositional Action Recognition}
As mentioned in~\Secref{sec:expr:comp_fewshot}, we also evaluate on the few-shot compositional action recognition task in~\cite{materzynska2019something}. For this setting, we use 88 \textit{base} action categories and 86 \textit{novel} action categories. We train on the base categories (113K/12K for training/validation) and fine-tune on few-shot samples from the novel categories (for 5-shot, 430/50K for training/validation; for 10-shot, 860/44K for training/validation). We use the same training recapie as in \ref{supp:impl:smthelse}.

\minisection{Optimization details} We train using $16$ frames with sample rate $4$ and batch-size $128$ (comprising $64$ videos and $64$ auxiliary synthetic datasets) on $8$ RTX 3090 GPUs. We train our network for 100 epochs with Adam optimizer~\cite{kingma2014adam} with a momentum of $9e-1$ and Gamma $1e-1$. Following~\cite{li2021improvedmvit}, we use $lr = 7e-5$ with half-period cosine decay. 

\minisection{Regularization details} We use weight decay of $1e-4$, and a dropout~\cite{Hinton2012dropout} of $5e-1$ before the final perdition.

\minisection{Training details} We use standard crop size of $224$, and we jitter scales from $256$ to $320$. 

\minisection{Inference details} We take 3 spatial crops per single clip to form predictions over a single video in testing.

\subsection{Something-Something v2}
\label{supp:impl:ssv2}
\minisection{Dataset} The SSv2~\cite{materzynska2019something} is a $\sim$160K-video dataset contains 174 action categories of common human-object interactions. We follow the official splits from~\cite{goyal2017something}.

\minisection{Optimization details} For the standard SSv2~\cite{materzynska2019something} dataset, we train using $16$ frames with sample rate $4$ and batch-size $128$ (comprising $64$ videos and $64$ auxiliary synthetic datasets) on $8$ RTX 3090 GPUs. We train our network for 100 epochs with Adam optimizer~\cite{kingma2014adam} with a momentum of $9e-1$ and Gamma $1e-1$. Following~\cite{li2021improvedmvit}, we use $lr = 7e-5$ with half-period cosine decay. 

\minisection{Regularization details} We use weight decay of $1e-4$, and a dropout~\cite{Hinton2012dropout} of $5e-1$ before the final classification.

\minisection{Training details} We use a standard crop size of $224$, and we jitter the scales from $256$ to $320$ along with RandomFlip.

\minisection{Inference details} We take 3 spatial crops per single clip to form predictions over a single video in testing as in ~\cite{li2021improvedmvit}.

\subsection{Ego4D}
\label{supp:impl:ego4d}

\minisection{Dataset} Ego4D~\cite{Ego4D2021} is a new large-scale dataset of more than 3,670 hours of video data, capturing the daily-life scenarios of more than 900 unique individuals from nine different countries around the world. The videos contain audio, 3D meshes of the environment, eye gaze, stereo and/or synchronized videos from multiple egocentric cameras.

\minisection{Metrics} In the Object State Change Temporal Localization task, the absolute error (in seconds) is used for evaluation. In the Object State Change Classification task, the top-1 accuracy is used for evaluation, following~\cite{Ego4D2021} protocol.


\minisection{Optimization details} We train using $16$ frames with sample rate $4$ and batch-size $128$ (comprising $64$ videos and $64$ auxiliary synthetic datasets) on $8$ RTX 3090 GPUs. We train our network for 10 epochs with Adam optimizer~\cite{kingma2014adam} with a momentum of $9e-1$ and Gamma $1e-1$. Following~\cite{li2021improvedmvit}, we use $lr = 1.5e-5$ with half-period cosine decay. Additionally, we used Automatic Mixed Precision, which is implemented by PyTorch.

\minisection{Training details} We use a standard crop size of $224$, and we jitter the scales from $256$ to $320$.  

\minisection{Inference details} We follow the official evaluation, both for the state change temporal localization and the state change classification tasks, available at \url{https://github.com/EGO4D/hands-and-objects}.

\subsection{AVA-2.2}
\label{supp:impl:ava}

\minisection{Dataset} AVA-2.2 (Atomic Visual Action) dataset~\cite{AVA2018} contains bounding box annotations for spatio-temporal localization of human actions. There are 211K training videos and 57K validation videos in the dataset. We report mean Average Precision (mAP) on 60 classes~\cite{AVA2018} on AVA v2.2 according to the standard evaluation protocol.

\minisection{Architecture} SlowFast~\cite{slowfast2019} and MViTv2~\cite{li2021improvedmvit} use a detection architecture with a RoI Align head on top of the spatio-temporal features. We follow their implementation to allow a direct comparison, elaborating on the RoI Align head proposed in SlowFast~\cite{slowfast2019}. First, we extract the feature maps from our {\smodel} model by using the RoIAlign layer. Next, we take the 2D proposal at a frame into a 3D RoI by replicating it along the temporal axis, followed by a temporal global average pooling. Then, we max-pooled the RoI features and fed them to an MLP classifier for prediction.

\minisection{Optimization details} To allow a direct comparison, we used the same configuration as in MViTv2~\cite{li2021improvedmvit}. We trained $16$ frames with sample rate $4$, depth of $16$ layers and batch-size $32$ (comprising $16$ videos and $16$ auxiliary synthetic datasets) on $8$ RTX 3090 GPUs. We train our network for 30 epochs with an SGD optimizer. We use $lr = 0.03$ with a weight decay of $1e-8$ together with early-stopping and a half-period cosine schedule of learning rate decaying.

\minisection{Training details} We use a standard crop size of $224$ and we jitter the scales from $256$ to $320$. We use the same ground-truth boxes and proposals that overlap with ground-truth boxes by $IoU > 0.9$ as in~\cite{slowfast2019}.

\minisection{Inference details} We perform inference on a single clip with $16$ frames. For each sample, the evaluation frame is centered in frame $8$. We take 1 spatial crop of $224$ with 10 different randomly sampled clips to aggregate predictions over a single video in testing.

\section{Additional Synthetic Datasets Details}
\label{supp:datasets}

\begin{figure*}[t!]
    \centering
    \includegraphics[width=0.9\linewidth]{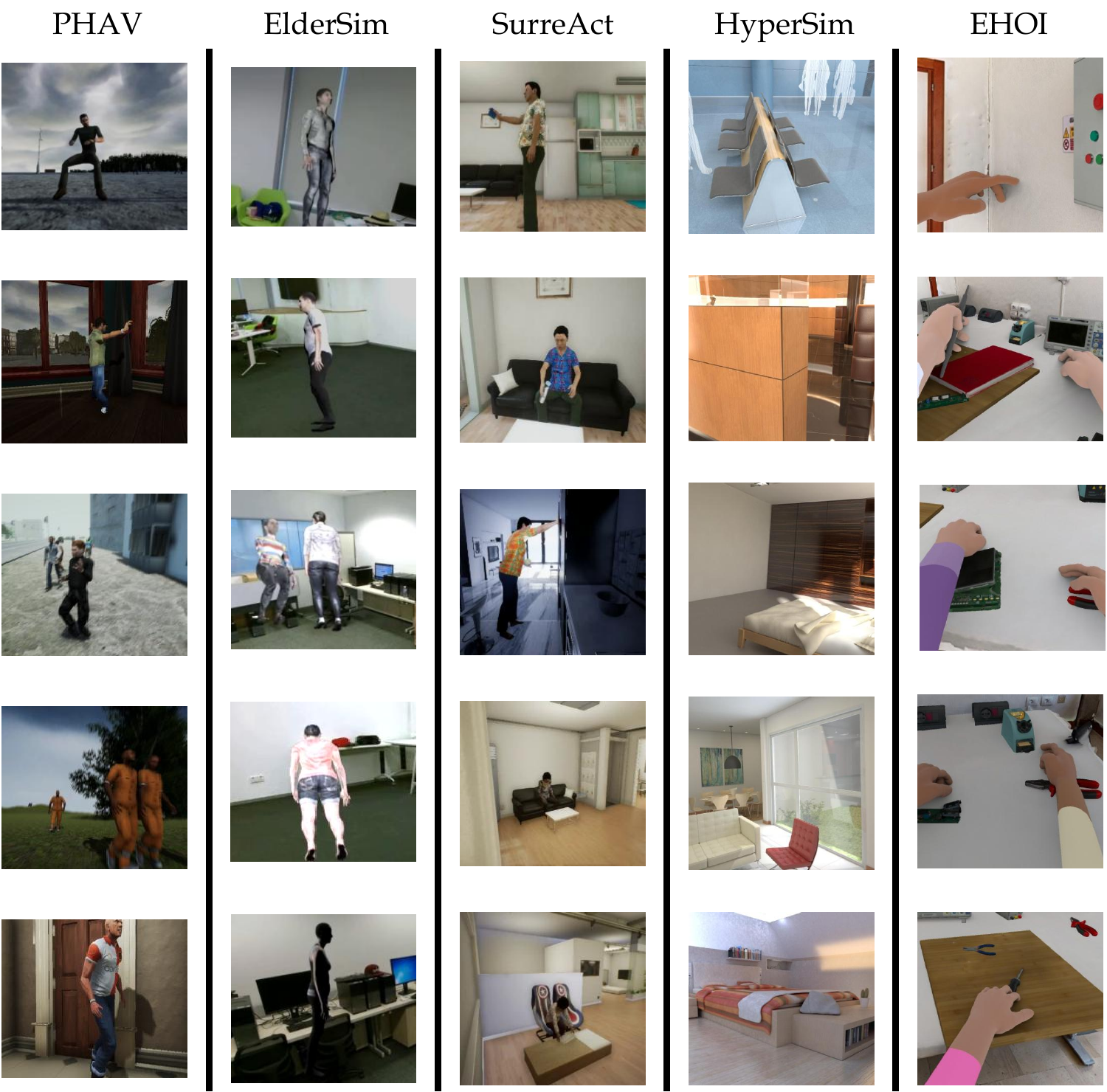}
    \captionof{figure}{\textbf{Synthetic Datasets Visualization.} Our training datasets for {\smodel} consist of several synthetic datasets that each emphasize different topics, including multi-views, static objects, hand-object interaction, and human motion activities.} 
    \label{supp:fig:data_vis}
\end{figure*}

Here we provide additional information about the ``auxiliary synthetic datasets'' (\Secref{supp:datasets:aux}), as well as the licenses and privacy policies for these datasets (\Secref{supp:datasets:Licenses}). \figgref{supp:fig:data_vis} shows examples of the synthetic videos we used to train on, while~\tabref{tab:dataset_details} presents the size of training samples across all synthetic and real datasets.


\setlength{\tabcolsep}{6pt}

\begin{table}[t]
    \begin{tabularx}
    {\linewidth}{@{}l c c c c}
        \toprule
        ~~\multirow{2}*{Dataset} & {Available} & {\#Training} & \multirow{2}*{Real/Synt.} \\
         & {Annots.} &  {Samples ($\times10^3$)} \\
        \midrule
        ~~PHAV          &  D+S   & 39.9  & Synt. \\
        ~~SURREACT      &  D+S+P3D & 108.3 & Synt. \\
        ~~ElderSim      &  P3D & 48.8  & Synt. \\
        ~~HyperSim      &  N+D & 31.1  & Synt. \\
        ~~EHOI          &  B+S & 20.0  & Synt. \\
        \midrule
        ~~SomethingElse   & - & 54.91    & Real \\
        ~~SSv2          & - & 157.4    & Real \\
        ~~AVA-2.2       & - & 193.3    & Real \\
        ~~Ego4D         & - & 41.1     & Real \\
        ~~Diving48      & - & 15.0     & Real \\
        \bottomrule
    \end{tabularx}
    
    \caption{\textbf{Real and synthetic dataset details}. We show (a) Top: the auxiliary synthetic datasets, and (b) Bottom: downstream real datasets. The available annotations are depth maps (D), segmentation (S), 3D poses (P3D), normal maps (N) and boxes (B).}
	\label{tab:dataset_details}
\end{table}

\subsection{Auxiliary Synthetic Datasets}
\label{supp:datasets:aux}

\minisection{Synthetic datasets} There has been recent interest in learning video understanding from synthetic data, including several popular synthetic datasets that have been proposed to improve video understanding. More specifically, a novel approach to data generation has been proposed by SURREACT~\cite{varol21_surreact} and UESTC~\cite{ji2019largescale} for synthesizing humans for actions. KIST SynADL~\cite{hwang2020eldersim} is a large-scale synthetic dataset of elders’ activities generated by the ElderSim engine~\cite{Hwang2020ElderSimAS}. The PHAV~\cite{DeSouza:Procedural:CVPR2017} dataset is a human action dataset that relies on the procedural generation of modern game engines. NTU RGB+D~\cite{shahroudy2016ntu} and UESTC RGB-D~\cite{ji2019largescale} are large-scale synthetic datasets that was proposed in order to allow the training of large video models for video understanding. HyperSim~\cite{roberts:2021} is a photo-realistic synthetic dataset for holistic indoor scene understanding. Egocentric Human-Object Interactions (EHOI)~\cite{leonardi2022egocentric} explores hand-object interaction in an industrial environment involving different objects, e.g. power supply, electrical panels, sockets, and more. In spite of the fact that these datasets contain different dataset styles, our approach is able to enhance video understanding models by utilizing synthetic data from various sources with multiple types of scene annotations. Next, we provide more details for each dataset separately.


\minisection{SURREACT~\cite{varol21_surreact}} The SURREACT dataset, which stands for Synthetic hUmans foR REal ACTions, renders video sequences from 3D skeleton joints by using a Skinned Multi-Person Linear Model (SMPL). The ground truth joints are extracted either by Kinect v2, or HMML~\cite{kanazawa2018learning}. SURREACT consists of \textbf{(1) NTU RGB+D}, which is a large-scale dataset for RGB-D human action recognition. It consists of 56,880 samples of 60 action classes collected from 40 subjects. The actions are generally categorized into three categories: 40 daily actions (e.g., drinking, eating, reading), nine health-related actions (e.g., sneezing, staggering, falling down), and 11 mutual actions (e.g., punching, kicking, hugging). These actions take place under 17 different scene conditions corresponding to 17 video sequences (i.e., S001–S017). The actions were captured using three cameras with different horizontal imaging viewpoints, namely, $-45^{\circ}$,$0^{\circ}$, and $+45^{\circ}$ degrees. Last, multi-modality information is provided for action characterization, including depth maps, 3D skeleton joint position, RGB frames, and infrared sequences; and \textbf{(2) UESTC RGB-D}, which contains 40 categories of aerobic exercise. The authors utilized two KinectV2 cameras in 8 fixed directions and 1 round direction to capture these actions with the data modalities of RGB video, 3D skeleton and depth map sequences.

\minisection{HyperSim~\cite{roberts:2021}} The HyperSim dataset is a high-resolution dataset consisting of 77,400 images from 461 indoor scenes with detailed per-pixel labels and corresponding ground truth geometry. It contains material and lighting information for every scene as well as dense per-pixel semantic instance segmentation, as well as complete camera information for every image. HyperSim was originally designed to handle the challenging per-pixel annotation of real data.

\minisection{KIST SynADL~\cite{hwang2020eldersim}} KIST SynADL is a synthetic dataset that focuses on elders' daily activities, which differ from other natural actions due to their high degree of variety. The activities of elders, such as \textit{sitting down} or \textit {washing face}, are more consistent psychically, shorter, and often rely on body position. Last, the dataset is generated using ElderSim~\cite{Hwang2020ElderSimAS}, a synthetic action simulation platform aimed at generating synthetic data on elders’ daily activities. Throughout the paper, we refer to ElderSim as KIST SynADL.

\minisection{Procedural Human Action Videos (PHAV)} The PHAV dataset is a diverse, realistic, and physically plausible dataset of human action videos. It contains a total of 39,982 videos, with more than 1,000 examples of each action in 35 categories across 7 different environments and 4 types of weather. The data is generated based on the existing motion-based real database CMU MOCAP database, for basic human animations. One of its key components is the use of Ragdoll physics to animate a human model while respecting basic physics properties such as connected joint limits, angular limits, weight, and strength. The videos are generated at 30fps and a resolution of 340x256.

\minisection{Egocentric Human-Object Interactions (EHOI)~\cite{leonardi2022egocentric}} EHOI is a synthetic image dataset that explores hand-object interaction in an industrial environment involving different objects, such as a power supply, electrical panels, sockets, etc. To create 3D models, several 3D scanners are applied, then using Blender, the authors generate the following: (1) photo-realistic RGB images; (2) depth maps; (3) semantic segmentation masks, objects, and hand-bounding boxes with contact states; and (4) distance between hands and objects in 3D space. The generated synthetic dataset contains a total of 20,000 images, 29,034 hands (of which 14,589 are involved in an interaction), 123,827 object instances (14,589 of which are active objects), and 19 object categories including portable industrial tools (e.g., screwdrivers, electrical boards) and instruments.


\subsection{Licenses and Privacy}
\label{supp:datasets:Licenses}
The license, PII, and consent details of each dataset are in the respective papers. In addition, we wish to emphasize that the datasets we use do not contain any harmful or offensive content, as many other papers in the field also use them. Thus, we do not anticipate a specific negative impact, but, as with any Machine Learning method, we recommend to exercise caution. 

\clearpage
\clearpage





\end{document}